\documentclass{ws-jdec}
\usepackage[T1]{fontenc}
\usepackage{times}
\renewcommand{\tifont}{\fontsize{14}{16}\selectfont\bfseries}
\renewcommand{\bibfont}{\fontsize{9}{11}\selectfont}
\usepackage[sort,compress,super]{cite}
\usepackage{bbold,bm,array,multirow}
\usepackage{float}
\usepackage{algpseudocode}
\usepackage[hidelinks,hypertexnames=false]{hyperref}
\hypersetup{
    pdftitle={Decomposing and Assessing Aleatoric and Epistemic Uncertainty in Wind Power Forecasting},
    pdfauthor={Yinsong Chen and Samson S. Yu}
}
\renewcommand{\wsdraftnote}{}
\floatstyle{ruled}
\newfloat{windalgorithm}{htbp}{loa}
\floatname{windalgorithm}{Algorithm}
\algnewcommand{\LineComment}[1]{\State \(\triangleright\) #1}

\begin{document}
\setlength{\pdfpagewidth}{8.5in}
\setlength{\pdfpageheight}{11in}
\raggedbottom
\catchline{}{}{}{}{}
\markboth{Y. Chen and S. S. Yu}{Uncertainty Decomposition in Wind Power Forecasting}
\title{Decomposing and Assessing Aleatoric and Epistemic Uncertainty in Wind Power Forecasting}
\author{Yinsong Chen$^a$ and Samson S. Yu$^a$}
\address{$^a$School of Engineering, Deakin University, Melbourne, Victoria 3216, Australia\\
E-mail: yinsong.chen@deakin.edu.au; samson.yu@deakin.edu.au}
\maketitle

\begin{abstract}
In wind power forecasting, aleatoric uncertainty (AU) describes the variability in power remaining for given inputs, while epistemic uncertainty (EU) reflects uncertainty about the model relating those inputs to power. We introduce Wind-UDA (Uncertainty Decomposition and Assessment), a framework for wind power forecasting. Proper scoring rules connect the entropy- and variance-based decompositions. The framework distinguishes the roles of training records and measurements available at prediction time. We evaluate the variance formulation using Bayes by Backprop, Monte Carlo DropConnect, and deep ensembles. In controlled synthetic experiments, sampled models could agree closely on an inaccurate mean prediction while overestimating outcome variability. Using real-world supervisory control and data acquisition (SCADA) records from an operating turbine, we compare ten-minute-ahead forecasts from power history with nowcasts incorporating target-time wind speed. With inputs fixed, increasing the training share from 10\% to 100\% more than doubled the deep ensemble's estimated mean AU. With the full training set fixed, adding current and recent wind speed to power history reduced it by about 63\%. Prediction error fell in both comparisons. Additional training records revise the estimated variability for given inputs, while a new measurement can explain part of the variability remaining at prediction time.
\end{abstract}


\begin{multicols}{2}
\raggedcolumns

\section{Introduction} 
Wind energy, a key renewable resource, has seen significant growth in installed capacity but poses challenges to grid reliability due to its intermittency.\cite{chen2022decomposition} Accurate forecasting is therefore critical for effective grid integration, while weather variability and power curve nonlinearities necessitate uncertainty quantification (UQ).\cite{10268627}

Wind power forecasting uncertainties arise from factors such as data quality and availability, measurement errors, modeling approaches, weather variability, and power curve biases.\cite{yan2022uncovering} These sources of uncertainty are broadly categorized as epistemic uncertainty (EU), associated with knowledge limitations that can be reduced with more comprehensive data or improved models, and aleatoric uncertainty (AU), which represents outcome variability remaining after conditioning on the forecasting inputs.\cite{caputo2023offshore}

Traditional UQ approaches in wind power forecasting typically yield a single total uncertainty (TU) estimate. Recent studies demonstrate that separating AU from EU improves forecasting performance and operational reliability.\cite{liu2023ultra,ding2022mixed,caputo2023offshore} Liu et al.~[\citen{liu2023ultra}] combined heteroscedastic regression for AU with Monte Carlo (MC) Dropout for EU, showing that their distinction mitigates the effects of observational noise and incomplete training data. Ding et al.~[\citen{ding2022mixed}] extended forecast-error UQ to explicitly include EU, demonstrating improved power system reliability. Beyond wind power, AU/EU disentanglement supports out-of-distribution (OOD) detection and active learning in broader machine learning settings.\cite{mukhoti2023deep,mindermann2022prioritized}

Two considerations motivate separate estimates of variability within a predictive model and uncertainty about the model itself. First, different UQ method families relate to AU, EU, and TU in distinct but partial ways. For example, probabilistic forecasting methods that fit a single predictive distribution, quantiles, or intervals typically quantify predictive uncertainty jointly.\cite{hullermeier2021aleatoric,10268627} Heteroscedastic regression explicitly models input-conditional noise variance as a proxy for AU, enabling data-driven estimation of irreducible noise, but is subject to known pitfalls including variance attenuation and mean--variance learning trade-offs.\cite{kendall2017uncertainties,seitzer2022pitfalls} Methods that sample network weights or combine independently trained networks represent model uncertainty through differences among their predictions.\cite{hullermeier2021aleatoric,lakshminarayanan2017simple} Empirical studies reveal interactions between the estimated components: Mucs\'{a}nyi et al.~[\citen{mucsanyi2024benchmarking}] benchmarked nineteen methods and found that decomposed AU and EU estimates remain highly correlated ($\text{rank corr.} \geq 0.78$) across nearly all evaluated approaches, and Valdenegro-Toro and Saromo Mori~[\citen{valdenegro2022deeper}] observed unintended interactions between the learning of AU and EU objectives, contrary to the assumed independence. These findings indicate that reliable disentanglement requires a principled framework beyond method selection alone.

Second, AU and EU convey operationally distinct information. AU reflects variability remaining given the forecasting inputs, including unresolved wind fluctuations, sensor noise, and power curve variability. EU reflects reducible model ignorance arising from sparse training coverage or distribution shift.\cite{kendall2017uncertainties,der2009aleatory} Aggregating both into a single TU obscures this distinction: EU drives active learning and OOD detection, while AU informs irreducible risk quantification and reliability assessment.\cite{mukhoti2023deep,mindermann2022prioritized}

Existing studies establish the value of AU/EU modeling for wind power prediction and identify uncertainty sources throughout the forecasting chain.\cite{liu2023ultra,yan2022uncovering} The remaining challenge is to connect the meaning of the two components, their estimation, and their evaluation across forecasting methods and wind operating conditions. In particular, additional forecasting inputs can change the conditional variability to be predicted, whereas additional training records refine knowledge of that conditional relationship. We introduce Wind-UDA (Uncertainty Decomposition and Assessment), a framework for wind power forecasting. It separates predictive uncertainty into aleatoric and epistemic components and examines how each responds to the information available during training and at the time of prediction.

The main contributions of this paper are threefold:
\begin{enumerate}
	\item A general decomposition framework is formulated using proper scoring rules, yielding entropy- and variance-based measures of AU and EU. Properties of AU and EU distinguish the roles of forecasting inputs and training information and characterize the posterior-consistency limit.
	\item A controlled evaluation protocol is developed to test the interpretation of AU and EU across forecasting methods. It examines how closely estimated AU tracks known outcome variability and shows how low EU can coexist with a shared error in the predicted mean.
	\item The proposed framework is applied to wind power nowcasting using supervisory control and data acquisition (SCADA) records from an operating turbine. Across all three investigated methods, models trained on more records showed closer agreement on mean power, while models given current and recent wind speed estimated less variability around their individual predictions.
\end{enumerate}

The remainder of this paper is organized as follows. Section \ref{sec:1} formulates the forecasting problem and presents the uncertainty decomposition and its variance-based implementation. Section \ref{sec:2} describes the assessment protocol and reports results from the synthetic experiments and the wind turbine SCADA study. Concluding remarks are given in Section \ref{sec:3}.

\section{Preliminaries and Methodologies}
\label{sec:1}
\subsection{Wind Power Forecasting as a Learning Problem} 
Wind power forecasting methods are broadly classified into persistence, physical, statistical, and hybrid methods.\cite{chen2022decomposition, yan2022uncovering, qian2019review} This study focuses on the statistical category, where deep learning models learn input--output mappings from historical data. We formulate wind power forecasting as a supervised learning problem.\cite{10268627} The learner is provided with a training dataset $\mathcal{D} = \{(x_i,y_i)\}_{i=1}^N \subset \mathcal{Z}$, where $\mathcal{Z} = \mathcal{X} \times \mathcal{Y}$ is the sample space. Each input $x_i$ comprises measurements available when the prediction is issued, and $y_i$ is the observed wind power at the target time. Given a hypothesis space $\mathcal{H}$, the objective is to induce a hypothesis $h \in \mathcal{H}$ that maps $\mathcal{X}$ to $\mathcal{Y}$ while minimizing the expected loss 
\begin{equation}
	L(h) = \int_{\mathcal{Z}} l(h,(x,y)) d \mathcal{P}(x,y),
\end{equation} where $\mathcal{P}$ denotes the unknown data-generating process and $l: \mathcal{H} \times \mathcal{Z} \rightarrow \mathbb{R}$ is a loss function. The hypothesis is typically selected by minimizing the empirical loss on the training data
\begin{equation}
	L_{\text{emp}}(h) = \sum_{i=1}^{N}l(h,(x_i,y_i)),  
\end{equation} resulting in the empirical loss minimizer $\hat{h} = \text{argmin}_{h \in \mathcal{H}} L_{\text{emp}}(h)$. This study considers scalar wind power targets and represents the hypothesis by a neural network (NN) with parameters $\theta$.\cite{chen5100162interplay}

The forecasting inputs $x$ and the training dataset $\mathcal{D}$ play distinct roles in uncertainty disentanglement. The inputs specify the information on which a prediction is conditioned, whereas the training data provide evidence for learning the conditional relationship. For example, additional wind measurements available at prediction time can explain power variability unresolved by power history alone. Additional training records with the same input variables improve knowledge of that relationship. AU is therefore defined relative to the supplied inputs, while EU concerns uncertainty about the learned predictive model.\cite{hullermeier2021aleatoric} This distinction connects the statistical decomposition to the physical information available in wind power forecasting.
\subsection{Uncertainty Representation}
\label{sec:uncertainty-representation}
Wind power forecasting is fundamentally a regression task, with a numerical output $y$. The conditional distribution $p(y|\theta,x)$ describes uncertainty in wind power given the input $x$ and model parameter $\theta$. This first-order probability distribution represents AU conditional on the specified model.\cite{hullermeier2021aleatoric}

EU is represented by a second-order probability distribution over the possible first-order predictive distributions. For a parameterized model, the learner's distribution $q(\theta|\mathcal{D})$ over model parameters induces this second-order distribution at each input $x$.\cite{hullermeier2021aleatoric} In a Bayesian neural network (BNN), $q$ denotes the exact or approximate posterior distribution over the network weights and biases.\cite{chen5100162interplay}

Marginalizing the first-order distributions over the second-order distribution yields the predictive distribution, written in density form as
\begin{equation}
	p(y|x, \mathcal{D}) = \int p(y|\theta,x) q(\theta|\mathcal{D})\,d\theta,
\end{equation} combining conditional outcome variability with uncertainty about the predictive model. The uncertainty measure introduced below quantifies TU from this distribution.

\noindent\begin{minipage}{\columnwidth}
\begin{figurehere}
	\centering
	\includegraphics[width=\columnwidth]{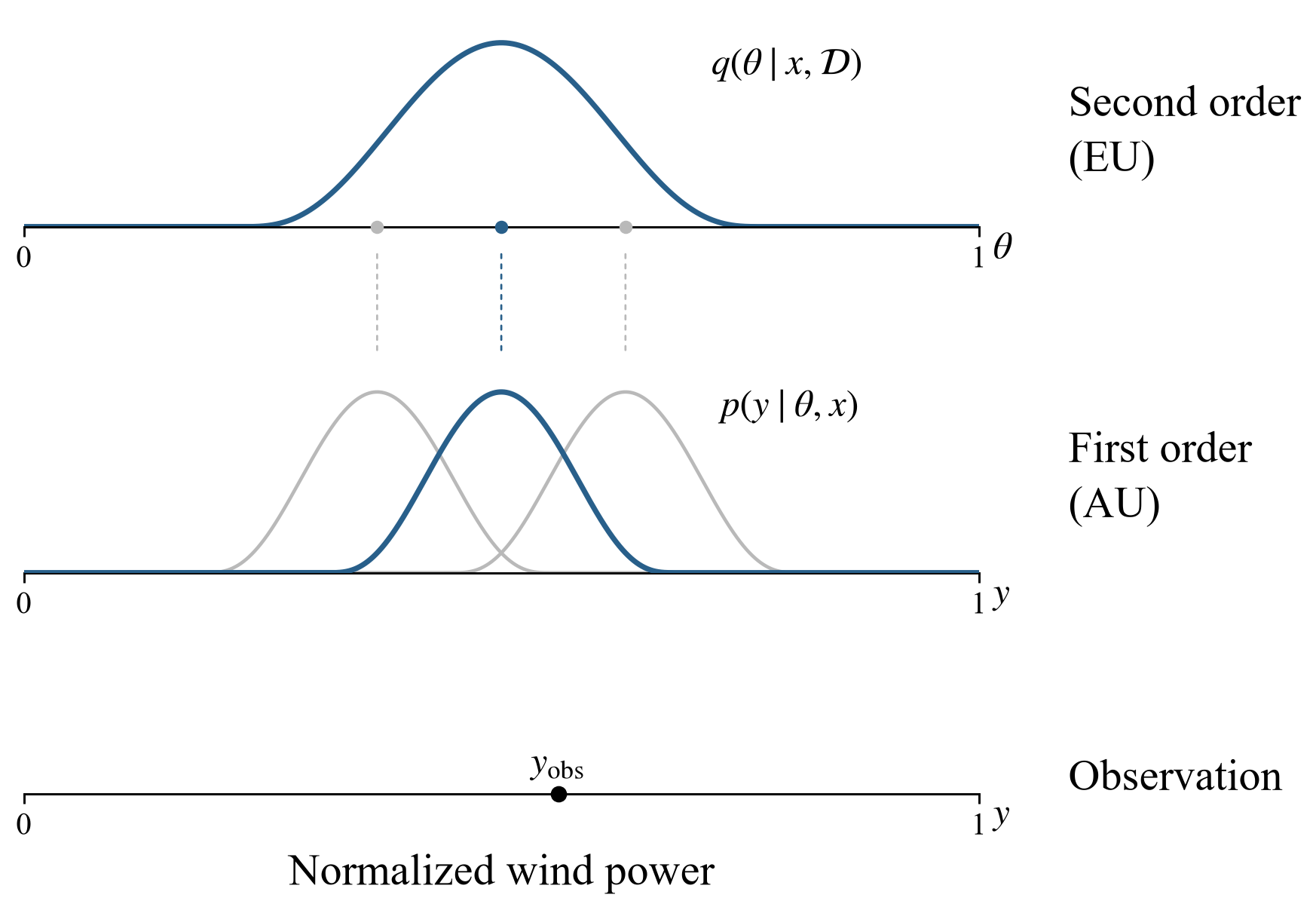}
	\caption{Two-level uncertainty representation.}
	\label{fig:uncertainty-hierarchy}
\end{figurehere}
\end{minipage}\par

Figure~\ref{fig:uncertainty-hierarchy} illustrates the hierarchical uncertainty representation at a fixed forecasting input $x$. Each curve in the middle panel is a first-order predictive distribution $p(y|\theta,x)$, whose spread represents AU conditional on a specified model. The upper density describes uncertainty about $\theta$ and thereby induces a second-order distribution over these predictive curves, representing EU. The marked parameter values and connecting lines link the two levels: in this example, varying the scalar location parameter $\theta$ shifts the predictive mean while preserving the conditional spread. The bottom panel shows an illustrative observed outcome $y_{\mathrm{obs}}$ on the wind-power axis. The following section quantifies these two forms of uncertainty through predictive loss.

\subsection{Uncertainty Disentanglement}
The two-level representation distinguishes variability in wind power conditional on a model from uncertainty about the model itself. To quantify both components on a common scale, we evaluate predictions using a proper scoring rule. The expected prediction loss, or risk, decomposes into Bayes risk and excess risk, represented by generalized entropy and divergence, respectively.\cite{gneiting2007strictly} Following the scoring-rule approach to regression uncertainty quantification,\cite{fishkov2025uncertainty} we apply this distinction to the two-level representation and then specialize the resulting decomposition to predictive variance.

\noindent\textbf{Definition 1 (Proper scoring rule).} A scoring rule $s(P,y)$ evaluates a predictive distribution $P$ against an observed outcome $y$, with smaller values indicating better predictions. If outcomes follow $Q$, the expected prediction loss is $E_{Y\sim Q}[s(P,Y)]$. The rule is proper if, for all distributions $P,Q$ in the considered class,\cite{10268627}
\begin{equation}
    E_{Y\sim Q}[s(Q,Y)] \leq E_{Y\sim Q}[s(P,Y)].
\end{equation}
The rule is strictly proper if equality holds only when $P=Q$.

For an outcome distribution $Q$, the Bayes risk is the minimum expected loss over the considered forecast class. Propriety ensures that forecasting $Q$ attains this minimum, which is the generalized entropy $H_s(Q)$. The excess risk of forecasting $P$ is its expected loss above the Bayes risk, represented by the associated divergence $d_s(P,Q)$:\cite{gneiting2007strictly,fishkov2025uncertainty}
\begin{equation}
    \begin{aligned}
        H_s(Q) &= \min_P E_{Y\sim Q}[s(P,Y)]\\
               &= E_{Y\sim Q}[s(Q,Y)],\\
        d_s(P,Q) &= E_{Y\sim Q}[s(P,Y)]-H_s(Q)\geq 0.
    \end{aligned}
    \label{eq:score-entropy-divergence}
\end{equation}
The prediction risk consequently decomposes as
\begin{equation}
    E_{Y\sim Q}[s(P,Y)]=H_s(Q)+d_s(P,Q).
    \label{eq:score-risk-split}
\end{equation}
Bayes risk measures the prediction loss remaining when the outcome distribution is known, while excess risk measures the penalty for using a different forecast.

We now apply this risk decomposition to the two-level predictive representation. For fixed $x$ and $\mathcal{D}$, write $P_\theta=p(\cdot|\theta,x)$, let $q$ denote the model parameter distribution conditional on $\mathcal{D}$, and set $\overline P=E_q[P_\theta]$. Each component distribution $P_\theta$ has Bayes risk $H_s(P_\theta)$. AU is the average of these component Bayes risks over $q$, while TU is the Bayes risk of the combined predictive distribution:
\begin{equation}
    U_T^s=H_s(\overline P),\qquad U_A^s=E_q[H_s(P_\theta)].
    \label{eq:score-tu-au}
\end{equation}
For outcomes following a component distribution $P_\theta$, issuing the combined forecast $\overline P$ incurs excess risk $d_s(\overline P,P_\theta)$. EU is this excess risk averaged over $q$. The following result relates these three quantities through the scoring-rule mixture decomposition.\cite{gneiting2007strictly,fishkov2025uncertainty}

\noindent\textbf{Theorem 1 (Score-based predictive uncertainty decomposition).} Let $q$ be a probability measure on the model parameter space and $\theta\mapsto P_\theta$ a measurable probability kernel on the outcome space. Let $s$ be a proper scoring loss on a class containing the component distributions and their mixture $\overline P$. Assume $(\theta,y)\mapsto s(P_\theta,y)$ and $(\theta,y)\mapsto s(\overline P,y)$ are measurable and absolutely integrable under $q(d\theta)P_\theta(dy)$. Then
\begin{equation}
    \begin{aligned}
        U_T^s &= U_A^s+U_E^s,\\
        U_E^s &:=E_q[d_s(\overline P,P_\theta)]\geq0.
    \end{aligned}
    \label{eq:score-decomposition}
\end{equation}
\noindent\textbf{Proof:} Applying Eq.~(\ref{eq:score-risk-split}) with $P=\overline P$ and $Q=P_\theta$, and averaging over $q$, gives
\begin{equation}
    \begin{aligned}
        H_s(\overline P)
        &=E_q E_{Y\sim P_\theta}[s(\overline P,Y)]\\
        &=E_q[H_s(P_\theta)]+E_q[d_s(\overline P,P_\theta)].
    \end{aligned}
\end{equation}
The first equality follows from the mixture definition and interchange of integrals; nonnegativity of EU follows from propriety.

For a strictly proper score, EU vanishes exactly when $P_\theta=\overline P$ for $q$-almost every $\theta$, meaning that the component models agree on the predictive distribution at the specified input.

\noindent\textbf{Information-theoretic interpretation.} Choosing the logarithmic loss $s_{\log}(P,y)=-\log p(y)$ makes the generalized entropy $H_s$ the differential entropy for continuous wind-power outcomes. Let $h$ denote differential entropy under the joint predictive model $\Theta\sim q$ and $Y|\Theta=\theta\sim P_\theta$, with $x$ and $\mathcal{D}$ fixed. TU is then the entropy $h(Y|x,\mathcal{D})$ of the predictive distribution. If $\theta$ were known, wind power would still vary according to $P_\theta$; its entropy $h(Y|\Theta=\theta,x,\mathcal{D})$ measures this remaining conditional uncertainty. Averaging the individual models' predictive entropies over $q$ gives AU.
\begin{equation}
    \begin{aligned}
        U_A^{\log}
        &=E_q[h(Y|\Theta=\theta,x,\mathcal{D})]\\
        &=h(Y|\Theta,x,\mathcal{D}).
    \end{aligned}
    \label{eq:score-conditional-entropy}
\end{equation}
EU therefore quantifies the expected reduction in predictive entropy from knowing the component model. This reduction is the mutual information between $Y$ and $\Theta$, conditional on the forecasting input and training data.\cite{depeweg2018decomposition,wimmer2023quantifying} Under Theorem~1,
\begin{equation}
    \begin{aligned}
        U_E^{\log}
        &=h(Y|x,\mathcal{D})-h(Y|\Theta,x,\mathcal{D})\\
        &=I_q(Y;\Theta|x,\mathcal{D})\\
        &=E_q[D_{\mathrm{KL}}(P_\theta\|\overline P)].
    \end{aligned}
    \label{eq:score-mi}
\end{equation}
Here $D_{\mathrm{KL}}$ denotes the Kullback--Leibler (KL) divergence. Equivalently, EU is the expected information gained about the model parameter from observing wind power at $x$. On a fixed output scale, differential entropy can be negative, while mutual information is nonnegative.

\noindent\textbf{Variance specialization for wind power regression.} For squared-error prediction, take $s_{\mathrm{sq}}(P,y)=(y-E_P[Y])^2$. The Bayes risk is then $H_{\mathrm{sq}}(Q)=\operatorname{Var}_Q(Y)$, and the excess risk is $d_{\mathrm{sq}}(P,Q)=(E_P[Y]-E_Q[Y])^2$. AU becomes the average conditional variance, and EU measures disagreement in the conditional means. This proper score identifies the predictive mean, with EU vanishing when the means agree $q$-almost surely. For example, a mixture assigning positive weight to both $\mathcal{N}(0,1)$ and $\mathcal{N}(0,4)$ has zero variance-based EU but positive logarithmic-score EU. This specialization underlies the two-head estimators used below.

With the squared-error scoring rule, the generalized entropy of the combined predictive distribution is its variance. Hence TU is
\begin{equation}
	U_T^{\mathrm{sq}}=H_{\mathrm{sq}}(\overline P)=\sigma^2(y|x,\mathcal{D}).
\end{equation}

\noindent\textbf{Standing assumptions.} For the variance decomposition, we condition on the observed training data $\mathcal{D}$ and assume the following:

\noindent\textit{(A1) Model distribution.} $q$ is a probability distribution over the model parameters $\theta$.

\noindent\textit{(A2) Finite predictive moments.} At every forecasting input $x\in\mathcal{X}$, the average conditional variance is finite, $E_q[\sigma^2(y|\theta,x)]<\infty$, and the variance of the conditional means is finite, $\sigma^2_q(\mu[y|\theta,x])<\infty$.

Theorem~1 then yields the predictive variance decomposition:

\noindent\textbf{Corollary 1 (Predictive variance decomposition).} Under (A1)--(A2), the total predictive variance decomposes as
\begin{equation}
	\label{eq:10}
	\sigma^2(y|x, \mathcal{D}) = \underbrace{\sigma^2_q(\mu[y|\theta,x])}_{\text{Epistemic Uncertainty}}+\underbrace{E_q[\sigma^2(y|\theta,x)]}_{\text{Aleatoric Uncertainty}}.
\end{equation}

\noindent\textbf{Proof:} Apply Theorem~1 with $s=s_{\mathrm{sq}}$, equivalently the law of total variance conditional on $x$ and $\mathcal{D}$.\cite{chung2000course}

Here $\mu[y|\theta,x]$ and $\sigma^2(y|\theta,x)$ are the mean and variance supplied by each predictive model. The choice of $q$ connects these components to the inference methods compared in Section~\ref{sec:2}.

The following properties distinguish information about the model from information supplied as forecasting inputs. They apply to the general scoring framework, with all expected scores assumed finite.

\noindent\textbf{Proposition 1 (EU and model information).} At fixed $x$ and $\mathcal{D}$, let $B$ denote additional evidence about the model, and let $q_B=q(\theta|\mathcal{D},B)$ be the conditional update of $q$ under the same joint model. Expectations over $B$ use the marginal distribution implied by this joint model. Assume $Y$ and $B$ are conditionally independent given $\theta,x,\mathcal{D}$, so that the component predictions $P_\theta$ remain fixed. Writing $\overline P_B=E_{q_B}[P_\theta]$, the expected reduction in EU is
\begin{equation}
    U_E^s(q)-E_B[U_E^s(q_B)]
    =E_B[d_s(\overline P,\overline P_B)]\geq0.
    \label{eq:eu-information}
\end{equation}
\noindent\textbf{Proof:} Conditional averaging gives $E_B[\overline P_B]=\overline P$ and $E_B[U_A^s(q_B)]=U_A^s(q)$. Applying Theorem~1 to the family $\overline P_B$ gives the stated identity. Thus, EU decreases on average as evidence refines knowledge of the model.

\noindent\textbf{Proposition 2 (AU and input information).} Let $P^*(Y|X)$ be the true conditional distribution of the outcome given the forecasting inputs. Let $A_s(X)=E_X[H_s(P^*(Y|X))]$ denote its average uncertainty, and define $A_s(X,Z)$ analogously for an additional input $Z$ available at prediction time. Then
\begin{equation}
    \begin{aligned}
        &A_s(X)-A_s(X,Z)\\
        &\quad=E_{X,Z}[d_s(P^*(Y|X),P^*(Y|X,Z))]\geq0.
    \end{aligned}
    \label{eq:au-information}
\end{equation}
\noindent\textbf{Proof:} Since $P^*(Y|X)=E_{Z|X}[P^*(Y|X,Z)]$, Theorem~1 applies conditionally on $X$; averaging over $X$ yields the result. For logarithmic scoring, the reduction is $I(Y;Z|X)$; for squared-error scoring, it is $E_X[\operatorname{Var}(E[Y|X,Z]|X)]$. This property characterizes the conditional uncertainty targeted by learned AU estimates.

\noindent\textbf{Corollary 2 (Posterior-consistency limit).} Fix $x$, write $P_x^*=P^*(\cdot|x)$, and let $q_N=q(\theta|\mathcal{D}_N)$. Suppose the model is well specified and the posterior concentrates at the true conditional distribution: for every $\varepsilon>0$,
\[
    q_N\!\left(\{\theta:\rho(P_\theta,P_x^*)>\varepsilon\}\right)
    \longrightarrow0\quad\text{almost surely},
\]
where $\rho$ is a metric on predictive distributions and almost-sure convergence is under the true training-data law.\cite{ghosal2017fundamentals} Assume $H_s(P)$ and $d_s(P_x^*,P)$ are continuous at $P_x^*$ in this metric and, evaluated at $P_\theta$, uniformly integrable under $q_N$ along almost every training-data sequence. Then, almost surely as $N\to\infty$,
\begin{equation}
    U_{E,N}^s\longrightarrow0,\qquad
    U_{A,N}^s,\,U_{T,N}^s\longrightarrow H_s(P_x^*).
    \label{eq:score-consistency}
\end{equation}
\noindent\textbf{Proof:} Posterior concentration, continuity, and uniform integrability give $U_{A,N}^s\to H_s(P_x^*)$ and $E_{q_N}[d_s(P_x^*,P_\theta)]\to0$. Since the combined forecast minimizes expected loss under its own distribution, $0\leq U_{E,N}^s\leq E_{q_N}[d_s(P_x^*,P_\theta)]$. Theorem~1 gives the TU limit.

The posterior-concentration condition follows, in particular, from $q_N\Rightarrow\delta_{\theta^*}$ almost surely when $P_{\theta^*}=P_x^*$ and $\theta\mapsto P_\theta$ is continuous at $\theta^*$ in $\rho$. For squared-error scoring, the limiting AU is the true conditional variance.

These properties motivate complementary evaluations: varying training information probes EU, while adding forecasting inputs probes conditional AU. A controlled variance-based noise comparison follows directly from Corollary~1: adding independent zero-mean noise of variance $\tau^2$ to every component adds $\tau^2$ to AU and leaves EU unchanged. The synthetic and SCADA training-data studies examine finite-sample responses relative to the posterior-consistency limit.

\subsection{Practical Implementation of Uncertainty Disentanglement for Wind Power Forecasting}
For a continuous model distribution $q$, the expectations in the variance decomposition can be approximated by Monte Carlo integration.\cite{chen5100162interplay} Given $S$ independent draws $\theta_s\sim q$, the estimators are
\begin{equation}
    \label{eq:12}
    \widehat U_{A,\mathrm{var}}=\frac{1}{S}\sum_{s=1}^S\sigma^2(y|\theta_s,x)
    \approx E_q[\sigma^2(y|\theta,x)],
\end{equation}
and
\begin{equation}
    \label{eq:13}
    \begin{split}
        \widehat U_{E,\mathrm{var}}
        &=\frac{1}{S}\sum_{s=1}^S\left(\mu[y|\theta_s,x]-\overline\mu_S(x)\right)^2,\\
        \overline\mu_S(x)&=\frac{1}{S}\sum_{s=1}^S\mu[y|\theta_s,x].
    \end{split}
\end{equation}
For regression visualization, the band $\overline\mu_S(x)\pm\sqrt{\widehat U_{A,\mathrm{var}}}$ represents one estimated aleatoric standard deviation about the predicted mean.

For a finite ensemble, set $q_S=S^{-1}\sum_{s=1}^S\delta_{\theta_s}$, where $\delta_{\theta_s}$ is a point mass at the fitted parameters of member $s$. The same sums are then the exact AU and EU components of this empirical mixture.

For independent draws from a fixed $q$, Eqs.~(\ref{eq:12}) and (\ref{eq:13}) converge almost surely to the corresponding components under (A2).

Bayes by Backprop (BBB) learns a distribution over network weights by variational inference,\cite{chen5100162interplay} while Monte Carlo DropConnect (MC-DropConnect) samples networks by masking connections. Deep ensembles use independently trained members.\cite{lakshminarayanan2017simple,fort2019deep} All three methods use Eqs.~(\ref{eq:12}) and (\ref{eq:13}); Algorithm~1 illustrates the deep-ensemble implementation.

The wind power point estimator is extended to a two-head estimator by incorporating a variance output head, $\sigma^2(y|x,\theta)$, commonly used in heteroscedastic regression tasks.\cite{le2005heteroscedastic,seitzer2022pitfalls} In our implementation, the two outputs parameterize $p(y|\theta,x)=\mathcal{N}(\mu[y|\theta,x],\sigma^2(y|\theta,x))$. Figure~\ref{fig:au} illustrates this architecture with distributions over network weights representing model uncertainty.

\noindent\begin{minipage}{\columnwidth}
\begin{figurehere}
	\centering
	\includegraphics[width=\columnwidth]{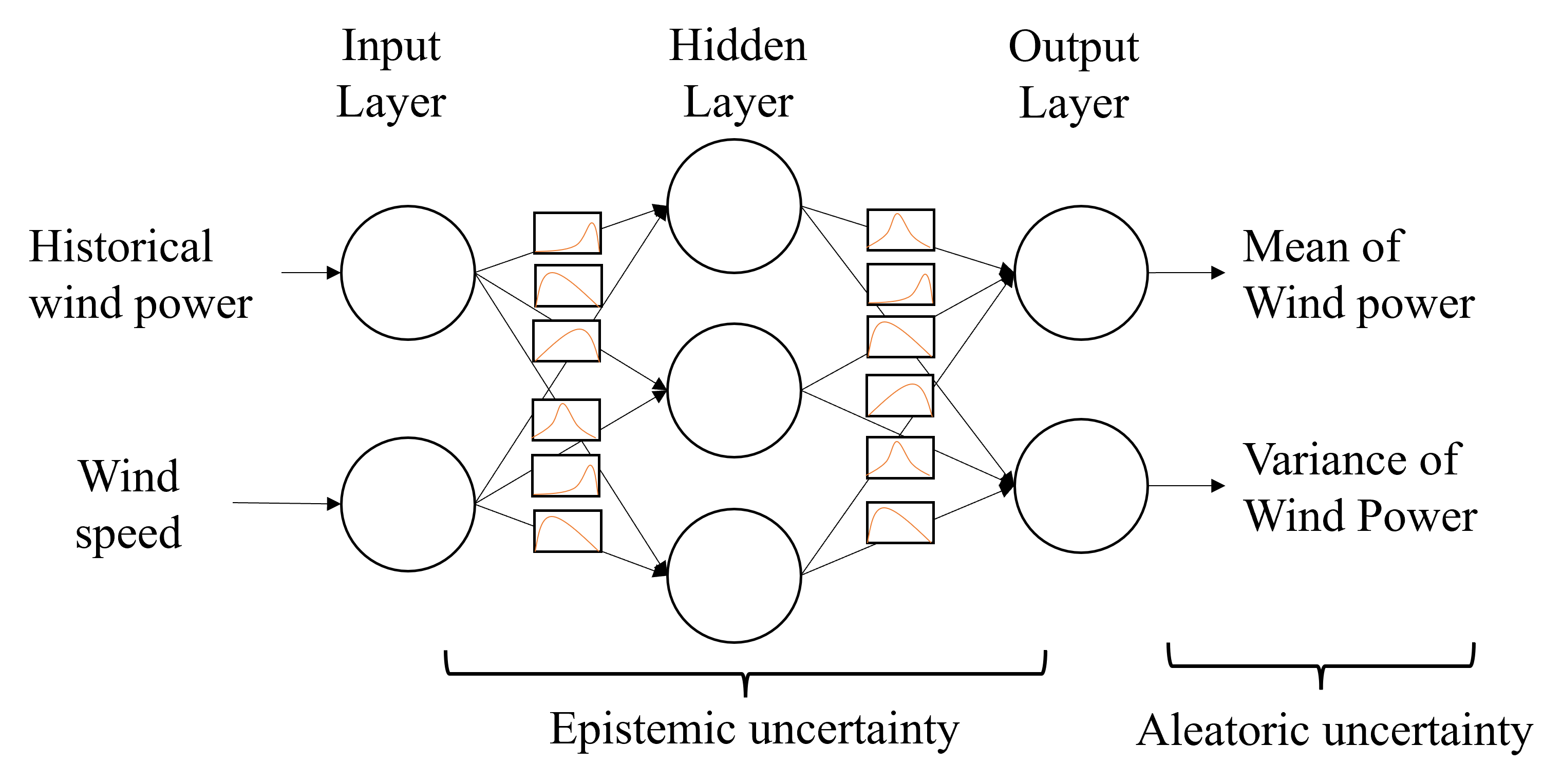}
	\caption{Two-head Bayesian neural predictor.}
	\label{fig:au}
\end{figurehere}
\end{minipage}\par

The scoring rule specifies how uncertainty is measured, while the training objective determines how the predictive distributions are learned. The Gaussian two-head predictor is trained using the negative log-likelihood (NLL) loss for the $i$-th sample:\cite{valdenegro2022deeper}
\begin{equation}
	l_{\text{NLL}}(x_i, \bar{y}_i, \theta_s) = \frac{\log \sigma^2(y_i|x_i, \theta_s)}{2} +\frac{(\mu(y_i|x_i, \theta_s)-\bar{y}_i)^2}{2\sigma^2(y_i|x_i, \theta_s)}.
\end{equation} Here, $\bar{y}_i$ denotes the observed label. To address variance underestimation during NLL training, the $\beta$-NLL loss has been proposed:\cite{seitzer2022pitfalls}
\begin{equation}
	l_{\beta\text{-NLL}}(x_i, \bar{y}_i, \theta_s) = \text{stop}(\sigma^{2\beta}(y_i|x_i, \theta_s))l_{\text{NLL}}(x_i, \bar{y}_i, \theta_s), 
\end{equation} where $\text{stop}(\cdot)$ preserves the value of the variance-dependent weight and treats it as constant during differentiation. At $\beta=0$, the loss recovers NLL; at $\beta=1$, the mean gradient has mean squared error (MSE) form while the variance head continues to learn.

In this way, the gradients of the empirical loss $L_{\beta\text{-NLL}} = \sum_{i=1}^N l_{\beta\text{-NLL}}(x_i, \bar{y}_i, \theta_s)$ w.r.t. $\mu$, $\sigma^2$ are
\begin{equation}
	\nabla_{\mu} L_{\beta\text{-NLL}}(\theta_s) = \sum_{i=1}^N \frac{\mu(y_i|x_i, \theta_s)-\bar{y}_i}{\sigma^{2-2\beta}(y_i|x_i, \theta_s)},
\end{equation}
\begin{equation}
	\nabla_{\sigma^2} L_{\beta\text{-NLL}}(\theta_s) = \sum_{i=1}^N \frac{\sigma^2(y_i|x_i, \theta_s)-(\bar{y}_i-\mu(y_i|x_i, \theta_s))^2}{2\sigma^{4-2\beta}(y_i|x_i, \theta_s)}, 
\end{equation} enabling standard gradient-based optimization.

\begin{windalgorithm}[H]
	\centering
	\caption{Uncertainty disentanglement for wind power forecasting via deep ensembles}\label{alg:ensemble}
	\begin{algorithmic}
		\LineComment{\textbf{Phase 1: Training ensembles of probabilistic models}}
		\For{s=1:S}
		\State Initialize parameters $\theta_s$ randomly.
		\State Minimize $L_{\beta\text{-NLL}}(\theta_s)$ w.r.t. $\theta_s$.
		\EndFor
		\LineComment{\textbf{Phase 2: Uncertainty disentanglement}}
		\State Given a test input $x^*$:
		\State \hspace{\algorithmicindent} Obtain $S$ predictions $\{(\mu_s, \sigma^2_s)\}_{s=1}^S$ from ensemble models.
		\State \hspace{\algorithmicindent} Compute epistemic uncertainty: 
		\[
		\text{EU}_{\text{var}} = \frac{1}{S} \sum_{s=1}^S \left(\mu_s - \frac{1}{S} \sum_{i=1}^S \mu_i \right)^2.
		\]
		\State \hspace{\algorithmicindent} Compute aleatoric uncertainty: 
		\[
		\text{AU}_{\text{var}} = \frac{1}{S} \sum_{s=1}^S \sigma^2_s.
		\]
	\end{algorithmic}
\end{windalgorithm} 

\textbf{Algorithm~\ref{alg:ensemble}} uses independently trained ensemble members. For stochastic inference methods, \textbf{Phase 1} instead fits the model distribution and draws $S$ parameter samples from it; the uncertainty calculation in \textbf{Phase 2} uses the same formulas.
 
\subsection{Model Misspecification and the Calibration Gap}
\label{sec:calibration-gap}
Corollary~2 connects the learned uncertainty components to the true conditional distribution under posterior consistency. Model misspecification can obstruct this connection, for example when a constant variance is used for heteroscedastic outcomes or the model class cannot represent the true conditional mean.\cite{chen5100162interplay}

The discrepancy from the true conditional distribution $p^*(y|x)$ can be examined through a Bregman decomposition.\cite{mucsanyi2024benchmarking} For $y\sim p^*(\cdot|x)$ and $\theta\sim q$ independent conditional on $x,\mathcal{D}$, the expected squared error decomposes as
\begin{equation}
	\begin{split}
	& E_{p^*(y|x)} E_q\!\left[(y-\mu[y|\theta,x])^2\right] \\
	& = \underbrace{\sigma^2_{p^*}(y|x)}_{\text{True Noise}} + \underbrace{\sigma^2_q(\mu[y|\theta,x])}_{\text{Epistemic Uncertainty}} + \underbrace{\Delta_{\text{bias}}(x)}_{\text{Model Bias}},
	\end{split}
\end{equation} where $\Delta_{\text{bias}}(x) = \left(E_q[\mu[y|\theta,x]] - E_{p^*}[y|x]\right)^2$ is the squared bias of the predictive mean.

The true conditional variance is the reference for assessing estimated AU. When a model misses the conditional mean, its expected squared residual includes the squared error in its mean prediction, which can inflate the variance it learns. For the AU estimate in Eq.~(\ref{eq:12}), the calibration gap is
\begin{equation}
	\Delta_{\text{cal}}(x) = E_q[\sigma^2(y|\theta,x)]
	- \sigma^2_{p^*}(y|x).
\end{equation} Positive and negative $\Delta_{\text{cal}}(x)$ indicate overestimation and underestimation of conditional variance, respectively.

Under the squared-error specialization of Corollary~2, both $\Delta_{\text{bias}}(x)$ and $\Delta_{\text{cal}}(x)$ vanish asymptotically.

Model capacity, training-data quantity, and optimization affect mean and variance estimation. This connection motivates examining fitted AU alongside EU in the dataset-size study in Section~\ref{sec:2}.
\section{Evaluation and Experiments}
\label{sec:2}
\subsection{Assessment Protocol}
Controlled comparisons are conducted to assess how AU and EU estimates respond to changes in forecasting information and modeling choices.\cite{wimmer2023quantifying,de2024disentangled,mucsanyi2024benchmarking} The synthetic experiments (T1--T3) use known conditional means and variances to assess estimation accuracy. The real-world experiments (R1--R2) use turbine SCADA records to evaluate forecasts against measured power and interpret AU and EU by operating condition. The experimental comparisons are organized around two questions about what determines the resulting AU and EU estimates:
\begin{itemize}
    \item \textbf{What information is available?} T1 and R1 compare training set sizes with the inputs fixed to examine estimation of the same conditional relationship. T2 and R2 hold the training cases fixed and vary the input information available for each prediction to examine how much variability an additional measurement can explain (Proposition~2).
    \item \textbf{How is uncertainty estimated?} All experiments compare BBB, MC-DropConnect, and deep ensembles as three ways of representing model uncertainty. With training data and inputs fixed, T3 pairs each method with three values of $\beta$ in the heteroscedastic regression loss to assess how both modeling choices affect the AU and EU estimates.
\end{itemize}

\subsection{Controlled Experiments on a Synthetic Sinusoidal Dataset}
\label{sec:toy}
The synthetic study uses a common data-generating model to distinguish changes in the conditional-variance target from changes in how accurately that target is learned. The three inference methods also reveal how fitted model disagreement responds to the same information changes.

\noindent\textbf{Data and experimental design.} Following the heteroscedastic sinusoidal setting in~[\citen{seitzer2022pitfalls,valdenegro2022deeper}], we generate
\begin{equation}
    Y=X\sin X+XZ+\epsilon,\qquad Z,\epsilon\overset{\mathrm{ind}}{\sim}\mathcal{N}(0,0.3^2),
    \label{eq:toy-generation}
\end{equation}
where $X$ is the scalar input, $Y$ is the output, and $Z$ is an additional variable that can be supplied to the predictor. The independent random variables $Z$ and $\epsilon$ are drawn for each sample, independently of $X$, with $\epsilon$ remaining unobserved. With $X$ alone, the true conditional mean and variance are $m_*(x)=x\sin x$ and $v_*(x)=0.09x^2+0.09$. When $Z$ is also observed, they become $m_*(x,z)=x\sin x+xz$ and $v_*(x,z)=0.09$. Thus, increasing the number of training observations retains the same conditional-variance target, whereas observing $Z$ changes that target.

We generate five synthetic datasets on the same uniform grid of 1000 training inputs in $[0,10]$, using a different random seed for each dataset to draw $Z$ and $\epsilon$ at each grid point. Within each dataset, a random permutation defines the nested training subsets used in T1. Validation and in-distribution test sets contain 256 and 512 grid points in $[0,10]$, respectively, with independent draws of $Z$ and $\epsilon$. We also evaluate the $X$-only models over $x\in(10,15]$ to examine extrapolation. Table~\ref{tab:toy-setup} summarizes the three experimental comparisons and model settings.

T2 uses the same 500 training cases $(x,z,y)$ from each dataset, supplying either $x$ alone or $(x,z)$ to the predictor. The same input choice applies to the shared validation and test cases. To keep the network dimensions fixed, both settings use a two-input network with $x$ as the first input. We fix the second input at zero when only $X$ is supplied and use the standardized value $z/0.3$ when $Z$ is included, where $0.3$ is the standard deviation of $Z$.

\end{multicols}
\noindent\begin{minipage}{\textwidth}
\begin{tablehere}
\tbl{Synthetic experimental design and model settings. All comparisons use the three methods across five datasets.\label{tab:toy-setup}}{%
\begin{minipage}{\linewidth}
    \centering
    \small
    \setlength{\tabcolsep}{5pt}
    \renewcommand{\arraystretch}{1.15}
    \textbf{(a) Experimental design}\par\smallskip
    \begin{tabular}{@{}p{0.23\textwidth}p{0.18\textwidth}p{0.16\textwidth}p{0.35\textwidth}@{}}
        \hline
        Experiment & Training size $N$ & Inputs & Comparison \\
        \hline
        T1: Training set size & 100, 300, 500, 1000 & $X$ & Nested training subsets \\
        T2: Input information & 500 & $X$ or $(X,Z)$ & Same cases, additional observed input \\
        T3: Method and loss weighting & 500 & $X$ & Three methods $\times$ $\beta\in\{0,0.5,1\}$; other settings fixed within each method \\
        \hline
    \end{tabular}
    \par\medskip
    \textbf{(b) Method settings}\par\smallskip
    \begin{tabular}{lccc}
        \hline
        Setting & BBB & MC-DropConnect & Deep ensembles \\
        \hline
        $\beta$ (T1--T2) & 0 & 0.4 & 0.5 \\
        Initial learning rate & 0.003 & 0.003 & 0.001 \\
        Epoch limit ($N=500$) & 1400 & 2800 & 1400 \\
        Decay epochs ($\times0.1$; $N=500$) & 800, 1200 & --- & --- \\
        Predictive samples / members & 30 & 30 & 5 \\
        Mask probability (hidden and output layers) & --- & 0.01 & --- \\
        \hline
    \end{tabular}
\end{minipage}}
\end{tablehere}
\end{minipage}\par
\begin{multicols}{2}
\raggedcolumns

\noindent\textbf{Models and training.} We compare how BBB, MC-DropConnect, and deep ensembles represent uncertainty about the conditional predictive distribution at a given input (Section~\ref{sec:uncertainty-representation}). BBB learns Gaussian distributions over network weights and biases by variational inference.\cite{chen5100162interplay} MC-DropConnect samples networks by randomly masking learned connections at prediction time.\cite{mobiny2021dropconnect} A deep ensemble assigns equal weight to networks trained on the same data from different initializations.\cite{lakshminarayanan2017simple} For each fitted model distribution, we estimate AU and EU using Eqs.~(\ref{eq:12}) and~(\ref{eq:13}) from the means and variances predicted by its weight samples, masks, or ensemble members, with counts given in Table~\ref{tab:toy-setup}. The predictive distribution is their equally weighted Gaussian mixture.

All networks have two 32-unit hidden layers with rectified linear unit (ReLU) activations, a linear mean head, and a softplus variance head. Training uses Adam with a batch size of 64. BBB uses a zero-mean Gaussian-mixture prior with standard deviations 1.5 and 0.1 and equal mixture weights; its KL coefficient is $0.1/B$, where $B$ is the number of minibatches per epoch.

Hyperparameters were selected through trial and error using validation continuous ranked probability score (CRPS) on two tuning seeds. In T1, epoch limits and decay epochs are scaled across training sizes to preserve the corresponding minibatch-update budgets within each method. Each final fit uses the checkpoint with the lowest validation CRPS of the predictive mixture.\cite{gneiting2007strictly}

\noindent\textbf{Evaluation metrics.} We use the following metrics to summarize AU and EU levels, assess the accuracy of the predicted mean and variance, and evaluate predictive quality. Specifically, mean AU and mean EU are the respective component estimates averaged over test inputs.

To assess estimation accuracy, we compare predictions with the known conditional mean and variance of the synthetic data. We use root mean squared error (RMSE) to compare the estimated AU $\hat v_i$ with the true conditional variance $v_i^*$ and mean-function MSE to compare the predictive mean $\hat m_i$ with the true conditional mean $m_i^*$. For $n$ test cases,
\begin{equation}
    \begin{aligned}
        \mathrm{RMSE}_{\mathrm{AU}}&=\sqrt{\frac{1}{n}\sum_{i=1}^n(\hat v_i-v_i^*)^2},\\
        \mathrm{MSE}_{\mathrm{mean}}&=\frac{1}{n}\sum_{i=1}^n(\hat m_i-m_i^*)^2.
    \end{aligned}
    \label{eq:toy-component-metrics}
\end{equation}
AU RMSE thus summarizes the magnitude of the variance calibration gap in Section~\ref{sec:calibration-gap} across test inputs.

Finally, CRPS evaluates the full predictive distribution against observed outcomes:\cite{gneiting2007strictly}
\begin{equation}
    \mathrm{CRPS}=\frac{1}{n}\sum_{i=1}^n\int_{-\infty}^{\infty}
    \left[F_i(t)-\mathbf{1}\{y_i\leq t\}\right]^2\,dt,
    \label{eq:toy-crps}
\end{equation}
where $F_i$ is the cumulative distribution function of the predictive Gaussian mixture, $y_i$ is the observed outcome, and $\mathbf{1}$ is the indicator function. Lower AU RMSE, mean-function MSE, and CRPS indicate better performance. Each metric is computed separately for each seed; tables report its mean and sample standard deviation across the five seeds.

The regression plots in Figs.~\ref{fig:toy-training-au} and~\ref{fig:toy-input-au} show seed 0. The fitted and true AU bands span one standard deviation around their respective means: $\overline\mu_S\pm\sqrt{\widehat U_{A,\mathrm{var}}}$ and $m_*\pm\sqrt{v_*}$. The model-disagreement and band-width comparisons in Figs.~\ref{fig:toy-training-eu} and~\ref{fig:toy-beta-au} use all five seeds.

\noindent\textbf{T1: Training set size.} Additional training observations help distinguish the shape of the mean curve from the noise around it. With 100 training observations, BBB underestimates the height of the last peak and predicts a wide AU band there (Fig.~\ref{fig:toy-training-au}). Even an observation close to the true peak then lies far from the fitted mean, so part of the apparent spread can arise from error in the mean curve. With 1000 observations, BBB recovers the peak and its band follows the true spread more closely. Across five datasets, its mean AU falls from 11.10 to 3.27, close to the true test-set average of 3.09 (Table~\ref{tab:toy-training}). MC-DropConnect and the deep ensemble already approximate the mean curve more closely at $N=100$, and both show smaller gains in AU accuracy.

\end{multicols}
\noindent\begin{minipage}{\textwidth}
\begin{figurehere}
    \centering
    \includegraphics[width=0.96\textwidth]{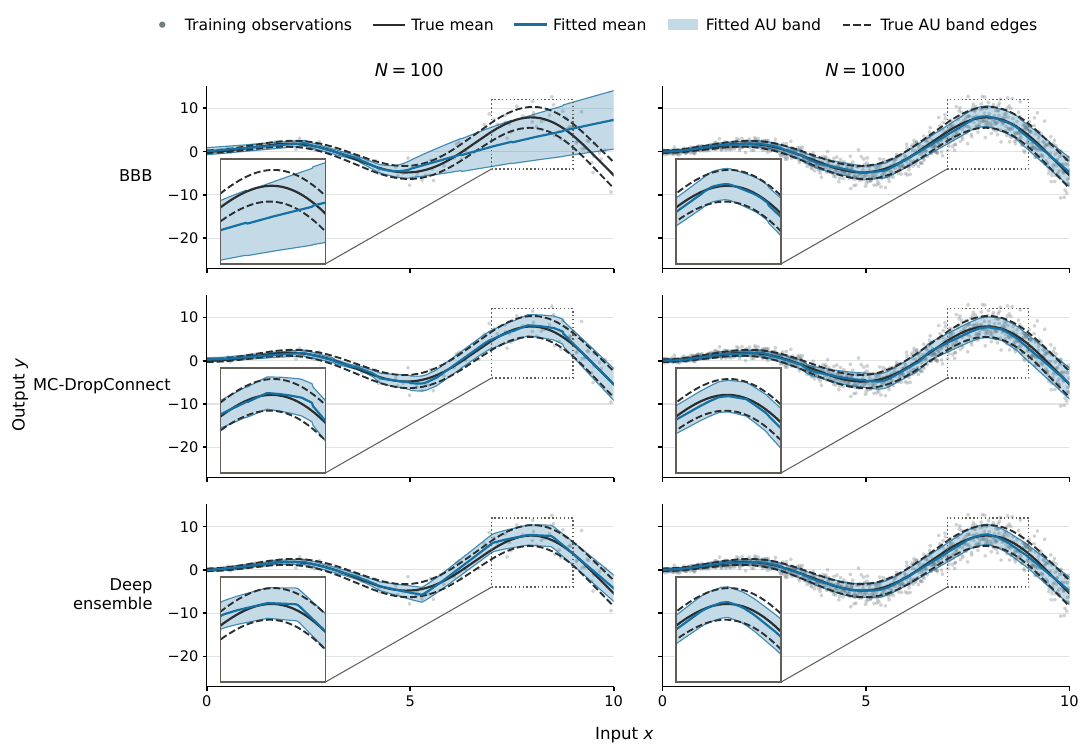}
    \caption{Regression fits with 100 and 1000 training observations from the same synthetic dataset (seed 0). The shaded band spans one predicted aleatoric standard deviation on either side of the fitted mean. Dashed curves give the corresponding true band. Each square inset enlarges the marked region around the last peak ($7\leq x\leq9$) using the same scale in all panels.}
    \label{fig:toy-training-au}
\end{figurehere}
\end{minipage}\par
\begin{multicols}{2}
\raggedcolumns

The missed peak also helps explain what EU measures. At $N=100$, BBB has the lowest mean EU of the three methods, yet the squared error of its predictive mean is about 20 times its mean EU (Table~\ref{tab:toy-training}). Most of the error is therefore shared by its sampled curves, as described by the decomposition in Section~\ref{sec:calibration-gap}. Increasing $N$ from 100 to 1000 reduces average disagreement for all three methods (Fig.~\ref{fig:toy-training-eu}), although the ensemble reaches its lowest EU at $N=500$.

At $N=1000$, the three methods achieve similar CRPS values (0.9028--0.9105), yet MC-DropConnect gives nearly four times the ensemble's mean EU. The ensemble also recovers the known variance curve most closely. Thus forecasts of similar overall quality can differ appreciably in how they represent uncertainty.

\end{multicols}
\noindent\begin{minipage}{\textwidth}
\begin{figurehere}
    \centering
    \includegraphics[width=\textwidth]{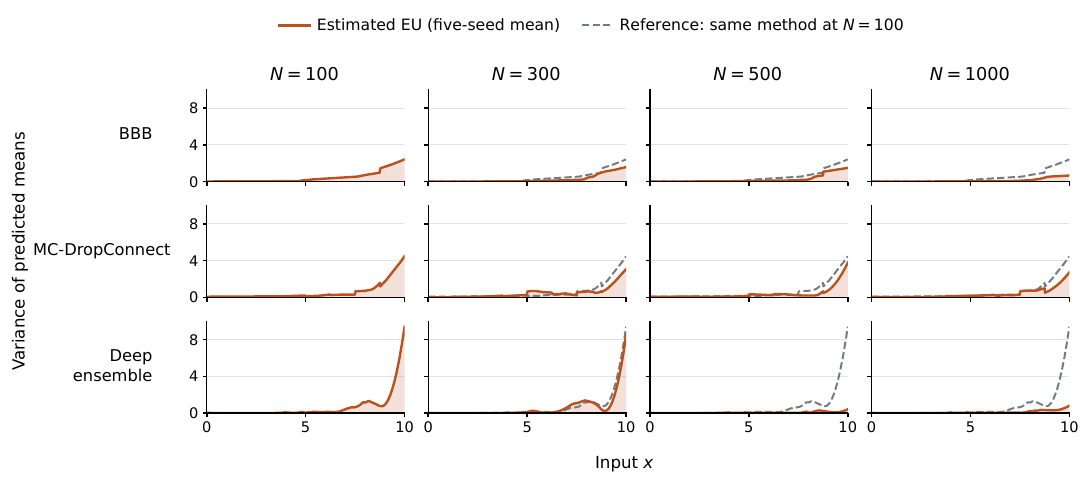}
    \caption{Estimated EU by training set size. At each $x$, the orange curve gives the variance of the models' predicted means, $\widehat U_{E,\mathrm{var}}$, averaged over five seeds. The dashed curve repeats the $N=100$ reference within each row.}
    \label{fig:toy-training-eu}
\end{figurehere}
\end{minipage}\par
\par\medskip
\noindent\begin{minipage}{\textwidth}
\begin{tablehere}
\tbl{Results by training set size: mean $\pm$ standard deviation across five seeds on the in-distribution test set. The true mean AU on this test set is 3.093.\label{tab:toy-training}}{%
\begin{minipage}{\linewidth}
\centering
\footnotesize
\setlength{\tabcolsep}{3pt}
\renewcommand{\arraystretch}{1.1}
\resizebox{\textwidth}{!}{%
\begin{tabular}{llccccc}
\hline
Method & $N$ & Mean AU & Mean EU & AU RMSE & Mean-function MSE & CRPS \\
\hline
BBB & 100 & $11.1002\pm 2.4560$ & $0.4408\pm 0.2004$ & $13.2177\pm 3.3295$ & $8.8824\pm 3.4767$ & $1.5031\pm 0.2285$ \\
BBB & 300 & $3.6717\pm 0.3832$ & $0.2597\pm 0.0873$ & $1.4531\pm 0.5611$ & $0.1907\pm 0.0763$ & $0.9188\pm 0.0205$ \\
BBB & 500 & $3.7558\pm 0.3182$ & $0.2456\pm 0.0425$ & $1.5344\pm 0.5987$ & $0.1817\pm 0.0678$ & $0.9106\pm 0.0136$ \\
BBB & 1000 & $3.2660\pm 0.0641$ & $0.1225\pm 0.0091$ & $0.4570\pm 0.0965$ & $0.0950\pm 0.0247$ & $0.9028\pm 0.0179$ \\
\hline
MC-DropConnect & 100 & $3.4977\pm 0.3328$ & $0.5637\pm 0.3582$ & $1.3949\pm 0.5456$ & $0.2895\pm 0.1617$ & $0.9425\pm 0.0347$ \\
MC-DropConnect & 300 & $3.5829\pm 0.5808$ & $0.4536\pm 0.1884$ & $0.9958\pm 0.9176$ & $0.1498\pm 0.0562$ & $0.9184\pm 0.0313$ \\
MC-DropConnect & 500 & $3.8465\pm 0.3784$ & $0.3649\pm 0.2010$ & $1.2464\pm 0.4266$ & $0.1827\pm 0.0547$ & $0.9277\pm 0.0263$ \\
MC-DropConnect & 1000 & $3.6192\pm 0.1547$ & $0.3682\pm 0.1607$ & $0.8770\pm 0.1476$ & $0.1170\pm 0.0400$ & $0.9105\pm 0.0195$ \\
\hline
Deep ensemble & 100 & $3.3275\pm 0.6204$ & $0.6340\pm 0.9174$ & $1.0811\pm 0.6199$ & $0.3254\pm 0.0711$ & $0.9447\pm 0.0205$ \\
Deep ensemble & 300 & $3.7141\pm 0.9290$ & $0.5594\pm 0.6400$ & $1.3132\pm 1.3415$ & $0.2579\pm 0.1967$ & $0.9221\pm 0.0265$ \\
Deep ensemble & 500 & $3.3170\pm 0.2645$ & $0.0481\pm 0.0191$ & $0.6025\pm 0.2769$ & $0.1114\pm 0.0374$ & $0.9065\pm 0.0172$ \\
Deep ensemble & 1000 & $3.1860\pm 0.0561$ & $0.0938\pm 0.0513$ & $0.3880\pm 0.1241$ & $0.1369\pm 0.0754$ & $0.9080\pm 0.0174$ \\
\hline
\end{tabular}}
\end{minipage}}
\end{tablehere}
\end{minipage}\par
\begin{multicols}{2}
\raggedcolumns

With all 1000 training observations in $[0,10]$, the three methods follow the true mean within that range but continue downward beyond $x=10$ as the true curve turns upward (Fig.~\ref{fig:toy-extrapolation}). For each method, the predicted means disagree more on average outside the training range, yet near the next peak the squared error of the predictive mean exceeds estimated EU. Over $(10,15]$, the central 90\% predictive intervals cover only 29.8--34.5\% of outcomes across the three methods, averaged over five seeds.

\end{multicols}
\noindent\begin{minipage}{\textwidth}
\begin{figurehere}
    \centering
    \includegraphics[width=\textwidth]{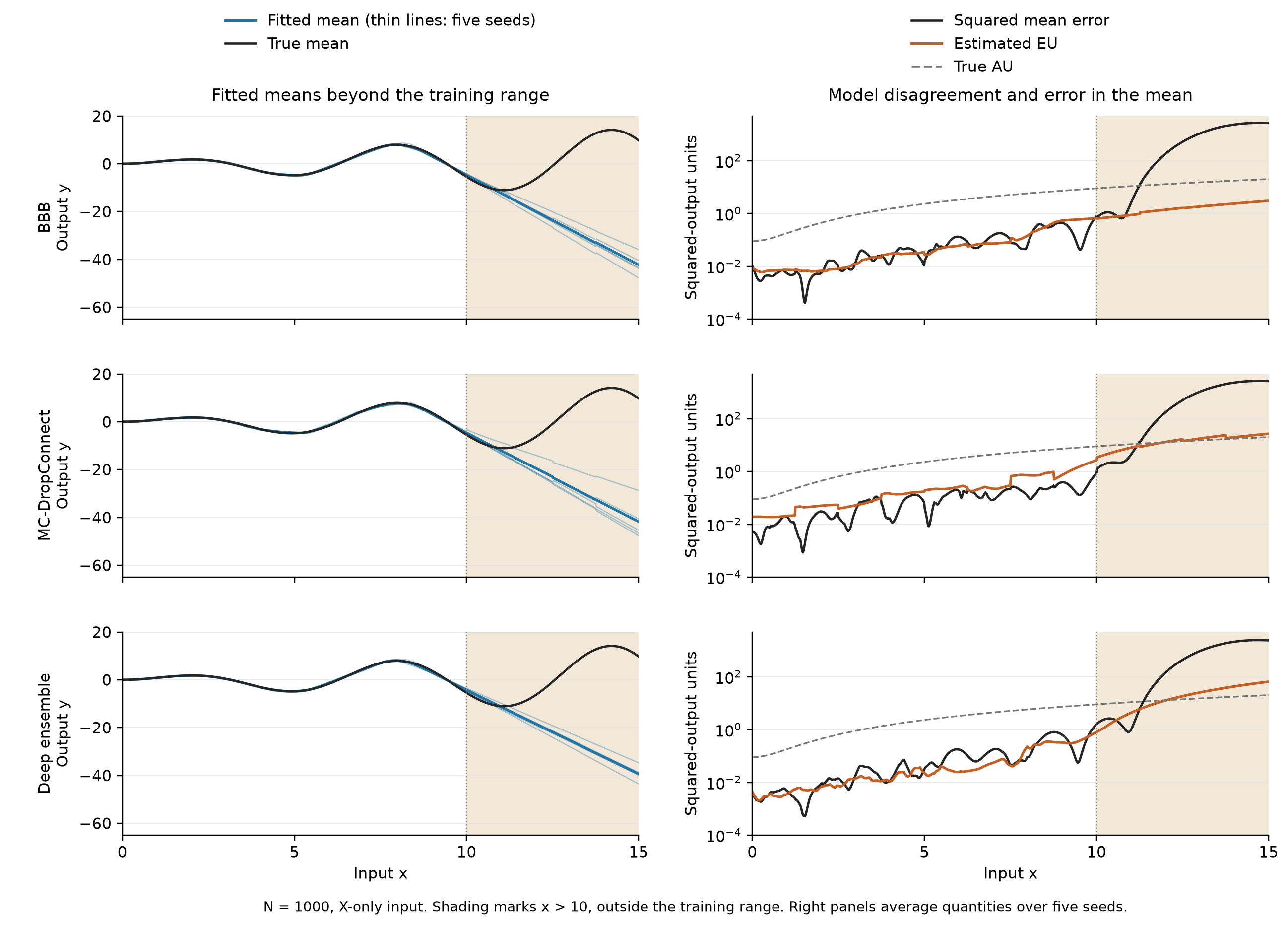}
    \caption{Extrapolation from 1000 training observations with $X$ as the only input. On the left, thin blue curves show the fitted means for five data seeds, the thick blue curve their average, and the black curve the true conditional mean. On the right, EU and the squared error of the predictive mean are each averaged over the five seeds and plotted alongside true AU on a logarithmic scale. Shading marks the extrapolation range $(10,15]$.}
    \label{fig:toy-extrapolation}
\end{figurehere}
\end{minipage}\par
\begin{multicols}{2}
\raggedcolumns

\noindent\textbf{T2: Input information.} T2 uses the same 500 training cases and varies whether $Z$ is available for each prediction. At a given $x$, the value of $Z$ shifts the expected outcome by $xZ$. With $X$ alone, this shift is unknown and its magnitude grows with $x$. This explains why the observations fan out in the left panels of Fig.~\ref{fig:toy-input-au}. When $Z$ is observed, the conditional mean accounts for each case's shift. The right panels show the same observations relative to this new mean, leaving only the noise $\epsilon$, with variance 0.090 throughout. Averaged over the test inputs, the true AU therefore falls from 3.093 to 0.090.

All three methods recover the direction predicted by Proposition~2. Adding $Z$ lowers estimated mean AU on each of the five datasets and also improves CRPS in every comparison (Table~\ref{tab:toy-input}). The methods differ in how closely they recover the remaining noise. The deep ensemble estimates a mean AU of 0.089 and has the lowest AU RMSE (0.039) across test inputs. BBB and MC-DropConnect retain higher mean AU estimates of 0.239 and 0.527, respectively. Their fitted bands remain visibly wider than the true band at large $x$ in Fig.~\ref{fig:toy-input-au}. Together, T1 and T2 illustrate two roles of information: additional training cases help estimate the spread around a given mean curve, while observing $Z$ helps predict why an individual outcome departs from that curve.

\end{multicols}
\noindent\begin{minipage}{\textwidth}
\begin{figurehere}
    \centering
    \includegraphics[width=0.99\textwidth]{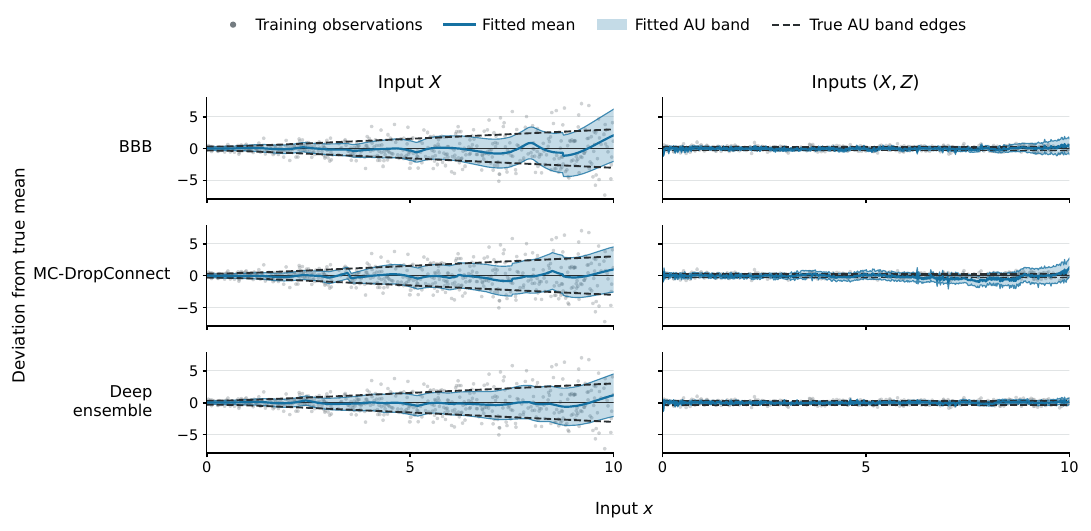}
    \caption{Variability remaining with $X$ alone and with $(X,Z)$ for the same 500 training observations (seed 0). Observations, fitted means, and AU bands are shown relative to the corresponding true conditional mean, $x\sin x$ on the left and $x\sin x+xz$ on the right. Dashed curves give the true one-standard-deviation band around zero.}
    \label{fig:toy-input-au}
\end{figurehere}
\end{minipage}\par
\par\medskip
\noindent\begin{minipage}{\textwidth}
\begin{tablehere}
\tbl{Results by forecasting input set at $N=500$: mean $\pm$ standard deviation across five seeds.\label{tab:toy-input}}{%
\begin{minipage}{\linewidth}
\centering
\footnotesize
\setlength{\tabcolsep}{3pt}
\renewcommand{\arraystretch}{1.1}
\resizebox{\textwidth}{!}{%
\begin{tabular}{llccccc}
\hline
Method & Inputs & Mean AU & Mean EU & AU RMSE & Mean-function MSE & CRPS \\
\hline
BBB & $X$ & $3.7558\pm 0.3182$ & $0.2456\pm 0.0425$ & $1.5344\pm 0.5987$ & $0.1817\pm 0.0678$ & $0.9106\pm 0.0136$ \\
BBB & $(X,Z)$ & $0.2386\pm 0.0531$ & $0.0700\pm 0.0187$ & $0.3983\pm 0.1699$ & $0.0527\pm 0.0211$ & $0.2148\pm 0.0134$ \\
\hline
MC-DropConnect & $X$ & $3.8465\pm 0.3784$ & $0.3649\pm 0.2010$ & $1.2464\pm 0.4266$ & $0.1827\pm 0.0547$ & $0.9277\pm 0.0263$ \\
MC-DropConnect & $(X,Z)$ & $0.5271\pm 0.0966$ & $0.2608\pm 0.0490$ & $0.7426\pm 0.2633$ & $0.1164\pm 0.0349$ & $0.2633\pm 0.0214$ \\
\hline
Deep ensemble & $X$ & $3.3170\pm 0.2645$ & $0.0481\pm 0.0191$ & $0.6025\pm 0.2769$ & $0.1114\pm 0.0374$ & $0.9065\pm 0.0172$ \\
Deep ensemble & $(X,Z)$ & $0.0890\pm 0.0051$ & $0.0228\pm 0.0085$ & $0.0387\pm 0.0127$ & $0.0209\pm 0.0036$ & $0.1860\pm 0.0070$ \\
\hline
\end{tabular}}
\end{minipage}}
\end{tablehere}
\end{minipage}\par
\begin{multicols}{2}
\raggedcolumns

\noindent\textbf{T3: Inference method and loss weighting.} T3 holds the 500 training cases and the input $X$ fixed and varies how prediction errors are weighted during training. Under Gaussian NLL ($\beta=0$), an observation assigned a larger variance has less influence on the mean update for the same prediction error. Increasing $\beta$ weakens this dependence. At $\beta=1$, equal errors receive equal weight in the mean update, while the variance is still learned.

Figure~\ref{fig:toy-beta-au} shows where this choice matters. Each row follows one method as $\beta$ increases from left to right. The dashed line marks zero error, where the estimated AU half-width equals the true half-width. In the bottom-left panel, the ensemble tracks the true width near $x=0$, but its band becomes too wide beyond $x=5$. At $x=10$, its estimated half-width averages about 4 across the five datasets, compared with a true value of about 3. At $\beta=0.5$, most of this excess disappears. Table~\ref{tab:toy-beta} shows the improvement across the full test range: the ensemble's AU RMSE falls from 3.703 to 0.603. BBB and MC-DropConnect also overestimate the width near $x=10$ under ordinary NLL. For both, the smallest AU RMSE occurs at $\beta=1$.

The effect on model disagreement depends on the inference method. For BBB and MC-DropConnect, increasing $\beta$ brings the predicted means closer together at each successive setting, although both methods achieve their lowest average CRPS at $\beta=0.5$. For the ensemble, increasing $\beta$ from 0.5 to 1 nearly doubles mean EU, from 0.048 to 0.096. This change is small relative to the mean AU of about 3.32 at both settings, so the total predictive variance changes little. The two ensemble forecasts also receive nearly identical CRPS values (0.9065 and 0.9060), even though their members disagree to different degrees.

\end{multicols}
\noindent\begin{minipage}{\textwidth}
\begin{figurehere}
    \centering
    \includegraphics[width=\textwidth]{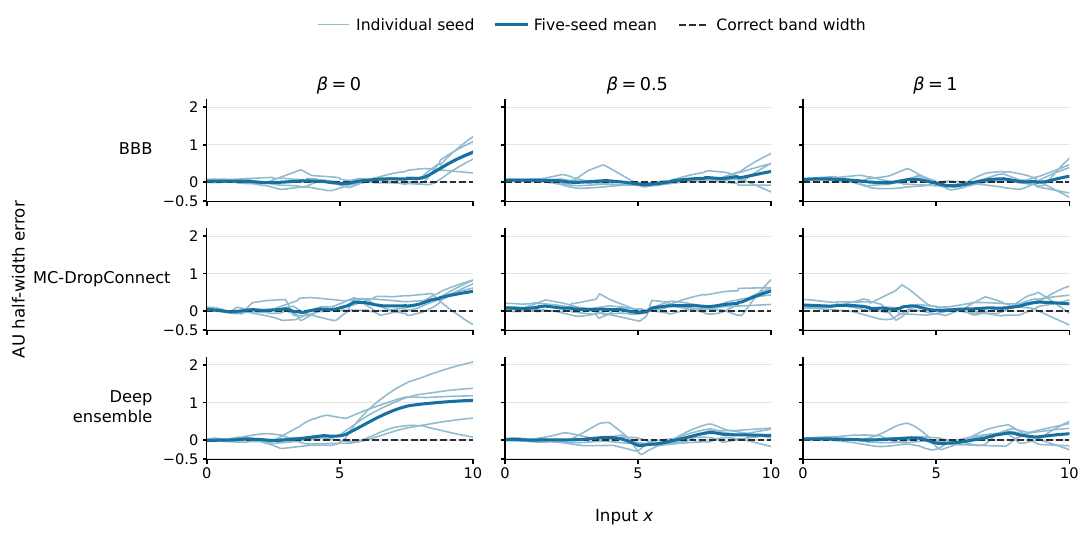}
    \caption{Effect of inference method and loss weighting on AU band width at $N=500$. Each curve shows fitted minus true half-width, $\sqrt{\widehat U_{A,\mathrm{var}}}-\sqrt{v_*}$, across the input range. Positive values indicate a band that is too wide and negative values one that is too narrow. Thin curves are individual seeds and the heavy curve is their mean.}
    \label{fig:toy-beta-au}
\end{figurehere}
\end{minipage}\par
\par\medskip
\noindent\begin{minipage}{\textwidth}
\begin{tablehere}
\tbl{Loss-weighting results at $N=500$ with $X$-only inputs: mean $\pm$ standard deviation across five seeds.\label{tab:toy-beta}}{%
\begin{minipage}{\linewidth}
\centering
\footnotesize
\setlength{\tabcolsep}{3pt}
\renewcommand{\arraystretch}{1.1}
\resizebox{\textwidth}{!}{%
\begin{tabular}{llccccc}
\hline
Method & $\beta$ & Mean AU & Mean EU & AU RMSE & Mean-function MSE & CRPS \\
\hline
BBB & 0.0 & $3.7558\pm 0.3182$ & $0.2456\pm 0.0425$ & $1.5344\pm 0.5987$ & $0.1817\pm 0.0678$ & $0.9106\pm 0.0136$ \\
BBB & 0.5 & $3.3756\pm 0.2048$ & $0.1286\pm 0.0262$ & $0.7527\pm 0.3907$ & $0.0862\pm 0.0169$ & $0.9026\pm 0.0216$ \\
BBB & 1.0 & $3.2114\pm 0.2022$ & $0.0963\pm 0.0174$ & $0.6624\pm 0.1278$ & $0.0734\pm 0.0267$ & $0.9067\pm 0.0243$ \\
\hline
MC-DropConnect & 0.0 & $3.8024\pm 0.2792$ & $0.4482\pm 0.2058$ & $1.4028\pm 0.2589$ & $0.1578\pm 0.0538$ & $0.9206\pm 0.0266$ \\
MC-DropConnect & 0.5 & $3.6611\pm 0.3070$ & $0.3395\pm 0.0933$ & $1.0751\pm 0.4826$ & $0.1213\pm 0.0540$ & $0.9129\pm 0.0249$ \\
MC-DropConnect & 1.0 & $3.5221\pm 0.3904$ & $0.2481\pm 0.1152$ & $0.8713\pm 0.3223$ & $0.1578\pm 0.0406$ & $0.9184\pm 0.0262$ \\
\hline
Deep ensemble & 0.0 & $5.4398\pm 1.6324$ & $1.0832\pm 0.7941$ & $3.7026\pm 2.5509$ & $1.5901\pm 1.4555$ & $1.0065\pm 0.1083$ \\
Deep ensemble & 0.5 & $3.3170\pm 0.2645$ & $0.0481\pm 0.0191$ & $0.6025\pm 0.2769$ & $0.1114\pm 0.0374$ & $0.9065\pm 0.0172$ \\
Deep ensemble & 1.0 & $3.3162\pm 0.2228$ & $0.0957\pm 0.0876$ & $0.7218\pm 0.1899$ & $0.1030\pm 0.0426$ & $0.9060\pm 0.0182$ \\
\hline
\end{tabular}}
\end{minipage}}
\end{tablehere}
\end{minipage}\par
\begin{multicols}{2}
\raggedcolumns

\subsection{Real-World Wind Turbine SCADA Study}
\label{sec:scada}

The SCADA study examines how training records and wind measurements affect estimates of turbine power and its uncertainty. R1 follows T1 by keeping the inputs fixed and using larger training subsets from the same period. T1 used a known noise curve to assess AU. Here, we interpret the estimates alongside forecast errors and interval coverage on later test data. R2 follows T2 by varying the inputs while keeping the training records fixed. We then examine where wind-speed measurements improve predictions along the turbine power curve.

\noindent\textbf{Data and experimental setup.} The public wind turbine SCADA dataset contains 50,530 records from 2018 at a nominal 10-minute resolution.\cite{scadadata} Figure~\ref{fig:scada} shows the wind speed--power relationship and the marginal distributions of both variables. Negative power is clipped to zero. The target is turbine power at time $t$. To compare all three input sets on the same observations, we retain only target times preceded by 36 consecutive 10-minute records (6 hours). Power and wind-speed measurements must be finite throughout this window and at the target time, with nonnegative wind speed. We split the original records chronologically into training, validation, and test periods in an 8:1:1 ratio. After window filtering, these periods contain 39,511, 4,945, and 4,979 eligible target times, respectively. Power is expressed in per unit (p.u.) by dividing by 3,600 kW, and wind speed is divided by 25 m/s.

\noindent\begin{minipage}{\columnwidth}
\begin{figurehere}
    \centering
    \includegraphics[width=0.95\columnwidth]{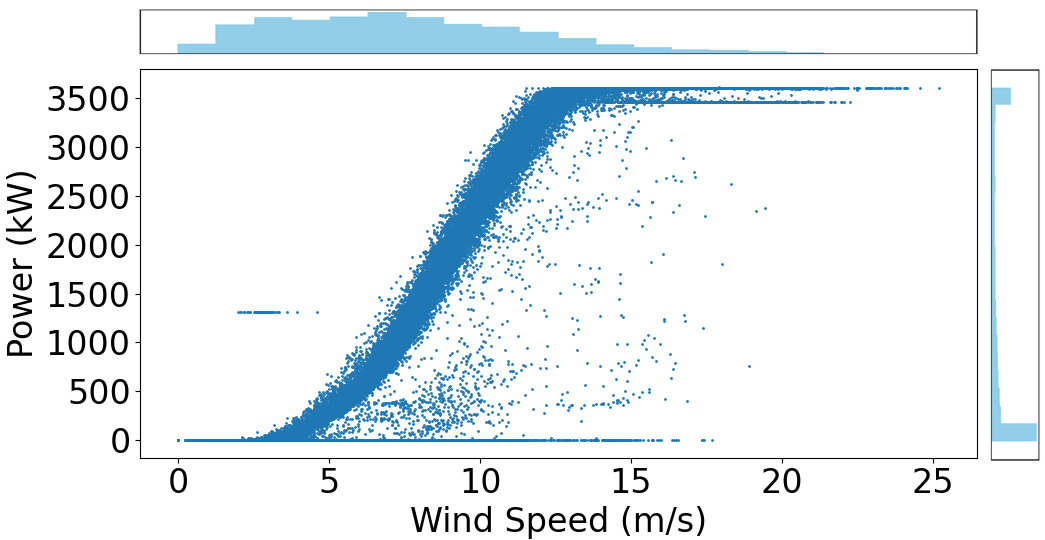}
    \caption{Wind speed and power in the SCADA dataset, with marginal histograms.}
    \label{fig:scada}
\end{figurehere}
\end{minipage}\par

Three input sets predict power at time $t$: \emph{Power: 100 min} uses $P_{t-1},\ldots,P_{t-10}$; \emph{Power: 6 h} uses $P_{t-1},\ldots,P_{t-36}$; and \emph{Power + speed} adds $V_t,V_{t-1},\ldots,V_{t-10}$ to the 100-minute power history. Here $P_t$ and $V_t$ denote measured power and wind speed. The \emph{Power + speed} setting uses wind speed measured at the target time to estimate power at that time, constituting nowcasting. To keep the network dimensions fixed, all three settings use a common input vector with 36 slots for past power, 36 for past wind speed, and one for current wind speed. Unused slots are set to zero.

R1 uses the \emph{Power + speed} input set with 10\%, 25\%, 50\%, and 100\% of the training records, corresponding to 3,951, 9,877, 19,755, and 39,511 samples. For each seed, a random permutation defines nested subsets spanning the training period. R2 uses all training records and compares the three input sets on the same validation and test targets. Each configuration is evaluated with five seeds.

All methods use three ReLU hidden layers of 64 units, a linear mean head, and a softplus variance head. Adam optimization uses a learning rate of 0.001, batch size 128, and reshuffling each epoch. Hyperparameters are selected using validation CRPS with two tuning seeds and then fixed for the five evaluation seeds. Table~\ref{tab:scada-settings} gives the method-specific settings. Within each method, the same epoch limit is used at every training share, so the maximum number of minibatch updates increases approximately tenfold from 10\% to 100\%. BBB uses the Gaussian-mixture prior described in the toy study, with a KL coefficient of $0.03/B$, where $B$ is the number of minibatches per epoch. The checkpoint with the lowest validation CRPS supplies the test predictions.

\noindent\begin{minipage}{\columnwidth}
\begin{tablehere}
\tbl{SCADA model settings. Epoch limits are maximum training budgets; checkpoints are selected by validation CRPS.\label{tab:scada-settings}}{%
\begin{minipage}{\linewidth}
    \centering
    \small
    \begin{tabular}{lrrr}
        \hline
         & Ensemble & MC-DropConnect & BBB \\
        \hline
        $\beta$ & 0.2 & 0.4 & 0 \\
        Epoch limit & 400 & 600 & 800 \\
        Members/MC draws & 5 & 30 & 30 \\
        Mask probability & --- & 0.01 & --- \\
        \hline
    \end{tabular}
\end{minipage}}
\end{tablehere}
\end{minipage}\par

Mean AU and EU use the variance estimators in Eqs.~(\ref{eq:12})--(\ref{eq:13}), averaged over test inputs. Predictive quality is assessed by CRPS, $\mathrm{RMSE}=\sqrt{n^{-1}\sum_i(y_i-\hat m_i)^2}$, and $\mathrm{NLL}=-n^{-1}\sum_i\log\overline p_i(y_i)$, where $\overline p_i$ is the predictive Gaussian mixture density. Its 0.05 and 0.95 quantiles define the central 90\% prediction interval; coverage is the fraction of targets inside that interval, and width is its average span. Metrics are computed per seed and summarized by their mean and sample standard deviation. Variance calibration is assessed by TU/MSE, the ratio of mean total predictive variance to mean squared prediction error, with both moments averaged across seeds before taking the ratio. Values above one indicate that predicted variance exceeds observed squared error.

\noindent\textbf{R1: Training set size.} As in T1, increasing the training share from 10\% to 100\% brings the predictions of mean power within each method closer together on average (Fig.~\ref{fig:scada-r1}(a)). All three methods also achieve lower RMSE and CRPS on the same test cases (Table~\ref{tab:scada-r1}). However, most of the gain in CRPS occurs by the 50\% share (Fig.~\ref{fig:scada-r1}(b)). Beyond that point, BBB's sampled models agree more closely still, with mean EU nearly halving, yet their combined forecast scores slightly worse.

BBB's estimated AU partly recovers after an initial drop, whereas its mean EU continues to fall. The training budget helps explain this sustained decline in disagreement. Whereas T1 matched the maximum number of updates across training sizes, R1 allows more updates on larger datasets under a fixed epoch limit. Even with the same 10\% of records, raising BBB's limit from 800 to 3200 epochs reduces its mean EU from $3.94\times10^{-4}$ to $8.30\times10^{-6}$~p.u.$^2$, close to the full-data value. CRPS improves from 0.0159 to 0.0106, but coverage of the nominal 90\% interval falls from 95.2\% to 85.5\% and NLL rises from $-2.17$ to 13.4. Thus further fitting alone can produce much closer agreement while leaving the predictive intervals too narrow.

The ensemble illustrates why closer agreement can also accompany wider intervals. Between 10\% and 100\%, its members agree more closely on mean power, but the variance they assign around their individual means more than doubles (Table~\ref{tab:scada-r1}). This rise in estimated AU outweighs the fall in EU. The ensemble's mean squared error also falls. Total predictive variance therefore rises from 0.65 to 1.65 times the observed mean squared error.

To separate the effect of additional records from that of further training, we fitted ensembles on the 10\% and 100\% training sets using the same initial weights for corresponding members. We compared the two fits after the same number of updates at each of two budgets (Table~\ref{tab:scada-r1-updates}). At both budgets, using the full training set increased mean AU in all five seeds. With the training set fixed, further updates reduced AU on average.

Each member learns its variance from the deviations of observed power around its fitted mean. With only 10\% of the records, further training roughly halved the members' mean squared error on the records they had seen, while their error on the remaining records from the same period changed little. As the training residuals shrank, the predictive intervals narrowed and covered only 86.7\% of test outcomes at the higher budget. At that same budget, including the remaining records gave the members a broader sample of deviations from which to learn. They predicted mean power more accurately on the test set while assigning a wider spread around it.

For MC-DropConnect, mean EU and CRPS change little between the 50\% and 100\% shares, but estimated AU rises by 38\%. At the full training share in Table~\ref{tab:scada-r1}, coverage exceeds the nominal 90\% for both methods, reaching 95.5\% for the ensemble and 96.9\% for MC-DropConnect.

\end{multicols}
\noindent\begin{minipage}{\textwidth}
\begin{figurehere}
    \centering
    \includegraphics[width=\textwidth]{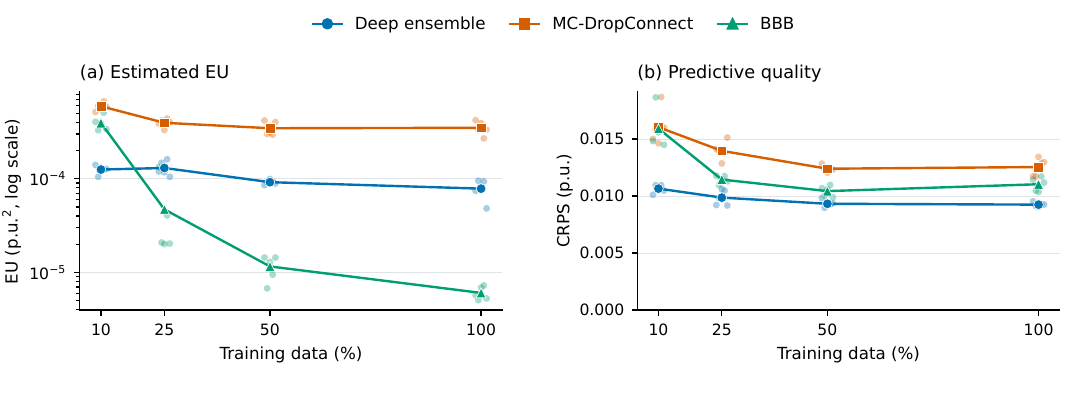}
    \caption{SCADA results by training set size. Panel (a) shows mean EU on a logarithmic scale. Panel (b) shows CRPS, with lower values indicating better forecasts. Large markers show five-seed averages and small dots show individual seeds.}
    \label{fig:scada-r1}
\end{figurehere}
\end{minipage}\par
\par\medskip
\noindent\begin{minipage}{\textwidth}
\begin{tablehere}
\tbl{Results by training set size with power and wind-speed inputs: mean $\pm$ sample standard deviation across five seeds. AU and EU are scaled by $10^{-3}$ and $10^{-4}$ p.u.$^2$, respectively. CRPS, RMSE, and interval width use p.u.; coverage is for the central 90\% predictive interval. TU/MSE is computed from seed-averaged moments.\label{tab:scada-r1}}{%
\begin{minipage}{\linewidth}
\centering
\footnotesize
\setlength{\tabcolsep}{2pt}
\renewcommand{\arraystretch}{1.08}
\begin{tabular}{@{}llrrrrrrrr@{}}
\hline
Method & \shortstack{Training\\data} & AU & EU & CRPS & RMSE & \shortstack{Coverage\\(\%)} & Width & NLL & \shortstack{TU/\\MSE} \\
\hline
\multirow{4}{*}{\shortstack[l]{Deep\\ensemble}} & 10\% & $0.487\pm 0.201$ & $1.254\pm 0.129$ & $0.0107\pm 0.0004$ & $0.0308\pm 0.0016$ & $92.1\pm 1.7$ & $0.0630\pm 0.0075$ & $-2.64\pm 0.08$ & $0.646$ \\
 & 25\% & $0.789\pm 0.190$ & $1.303\pm 0.235$ & $0.0099\pm 0.0007$ & $0.0290\pm 0.0006$ & $93.2\pm 0.8$ & $0.0644\pm 0.0081$ & $-2.79\pm 0.25$ & $1.089$ \\
 & 50\% & $0.801\pm 0.196$ & $0.919\pm 0.051$ & $0.0093\pm 0.0003$ & $0.0281\pm 0.0007$ & $93.9\pm 0.5$ & $0.0647\pm 0.0053$ & $-2.95\pm 0.24$ & $1.128$ \\
 & 100\% & $1.162\pm 0.177$ & $0.784\pm 0.190$ & $0.0092\pm 0.0002$ & $0.0275\pm 0.0003$ & $95.5\pm 0.3$ & $0.0738\pm 0.0034$ & $-3.21\pm 0.07$ & $1.646$ \\
\hline
\multirow{4}{*}{\shortstack[l]{MC-\\DropConnect}} & 10\% & $1.407\pm 0.337$ & $5.936\pm 0.545$ & $0.0160\pm 0.0016$ & $0.0387\pm 0.0037$ & $95.5\pm 1.7$ & $0.1199\pm 0.0114$ & $-2.31\pm 0.07$ & $1.329$ \\
 & 25\% & $1.291\pm 0.253$ & $3.955\pm 0.406$ & $0.0140\pm 0.0008$ & $0.0357\pm 0.0018$ & $95.8\pm 1.2$ & $0.1045\pm 0.0081$ & $-2.51\pm 0.08$ & $1.320$ \\
 & 50\% & $1.250\pm 0.230$ & $3.460\pm 0.596$ & $0.0124\pm 0.0003$ & $0.0332\pm 0.0013$ & $96.1\pm 0.9$ & $0.0990\pm 0.0067$ & $-2.70\pm 0.10$ & $1.449$ \\
 & 100\% & $1.722\pm 0.519$ & $3.497\pm 0.593$ & $0.0125\pm 0.0008$ & $0.0331\pm 0.0028$ & $96.9\pm 0.7$ & $0.1080\pm 0.0107$ & $-2.75\pm 0.08$ & $1.885$ \\
\hline
\multirow{4}{*}{BBB} & 10\% & $1.732\pm 0.650$ & $3.941\pm 0.709$ & $0.0159\pm 0.0016$ & $0.0380\pm 0.0035$ & $95.2\pm 1.9$ & $0.1161\pm 0.0171$ & $-2.17\pm 0.52$ & $1.465$ \\
 & 25\% & $0.968\pm 0.203$ & $0.471\pm 0.492$ & $0.0114\pm 0.0003$ & $0.0325\pm 0.0014$ & $89.7\pm 3.0$ & $0.0670\pm 0.0121$ & $0.42\pm 6.17$ & $0.958$ \\
 & 50\% & $1.247\pm 0.249$ & $0.116\pm 0.034$ & $0.0104\pm 0.0005$ & $0.0305\pm 0.0015$ & $91.4\pm 0.9$ & $0.0640\pm 0.0051$ & $-2.07\pm 2.38$ & $1.350$ \\
 & 100\% & $1.268\pm 0.272$ & $0.060\pm 0.010$ & $0.0110\pm 0.0006$ & $0.0303\pm 0.0006$ & $91.4\pm 1.4$ & $0.0720\pm 0.0078$ & $-3.00\pm 0.10$ & $1.390$ \\
\hline
\end{tabular}
\end{minipage}}
\end{tablehere}
\end{minipage}\par
\par\medskip
\noindent\begin{minipage}{\textwidth}
\begin{tablehere}
\tbl{Deep ensemble estimates of mean test-set AU at matched training budgets ($10^{-3}$~p.u.$^2$).\label{tab:scada-r1-updates}}{%
\begin{minipage}{\linewidth}
    \centering
    \small
    \begin{tabular}{@{}lrrr@{}}
        \hline
        Updates per member & 10\% of records & 100\% of records & Paired difference ($100\%-10\%$) \\
        \hline
        Low (9,300--12,400) & $0.544\pm0.286$ & $1.291\pm0.185$ & $+0.747\pm0.459$ \\
        High (61,800--123,600) & $0.295\pm0.182$ & $1.179\pm0.260$ & $+0.885\pm0.369$ \\
        \hline
    \end{tabular}
    \par\smallskip\raggedright\footnotesize
    Values are means $\pm$ sample standard deviations across five seeds. For each seed, the lower and higher budgets were the numbers of optimizer updates at the validation-selected checkpoints in the original 10\% and 100\% runs, respectively. At each update count, we measured AU on the same test set for ensembles trained on 10\% and 100\% of the records.
\end{minipage}}
\end{tablehere}
\end{minipage}\par
\begin{multicols}{2}
\raggedcolumns

\noindent\textbf{R2: Input information.} Starting from 100 minutes of past power measurements, we compare two ways of adding information while keeping the training records and test cases fixed (Table~\ref{tab:scada-r2}).
\begin{itemize}
    \item \textbf{Longer power history.} Extending the power history from 100 minutes to six hours changes mean CRPS by less than 3\% for each method. Mean AU is nearly unchanged for the ensemble and rises for MC-DropConnect and BBB.
    \item \textbf{Additional wind-speed measurements.} The power-only models predict turbine power ten minutes ahead. For the nowcast, we retain the 100-minute power history and add wind speeds measured over that history and at the target time. Mean AU falls by 48.4--62.6\% and CRPS by 45.6--57.5\% across the three methods. Squared prediction error also falls in each of the five runs for every method.
\end{itemize}

Earlier power measurements add little predictive value here, whereas wind speed helps explain variation in current power that recent power history alone leaves unresolved. This is the role of an informative input illustrated in T2.

Although all three methods estimate lower AU when wind speed is available, their EU responds differently. The ensemble and BBB show closer agreement among their predicted means. For MC-DropConnect, however, mean EU rises even as CRPS improves. Wind measurements therefore improve the combined forecast across all three methods, while their effect on model disagreement depends on the method.

Table~\ref{tab:scada-r2} shows how the narrower forecasts perform against measured power. For the ensemble, adding wind speed nearly halves the average width of the nominal 90\% interval, from 0.1509 to 0.0738 p.u., while coverage remains at 95.5\%. Across all three methods, mean squared error falls faster than total predictive variance. Consequently, the forecasts become sharper, yet their mean total variance remains 1.390--1.885 times the observed mean squared error.

\end{multicols}
\noindent\begin{minipage}{\textwidth}
\begin{tablehere}
\tbl{Results by forecasting input set with the full training set: mean $\pm$ sample standard deviation across five seeds. AU and EU are scaled by $10^{-3}$ and $10^{-4}$ p.u.$^2$, respectively. CRPS, RMSE, and interval width use p.u.; coverage is for the central 90\% predictive interval. TU/MSE is computed from seed-averaged moments.\label{tab:scada-r2}}{%
\begin{minipage}{\linewidth}
\centering
\footnotesize
\setlength{\tabcolsep}{2pt}
\renewcommand{\arraystretch}{1.08}
\resizebox{\linewidth}{!}{%
\begin{tabular}{@{}llrrrrrrrr@{}}
\hline
Method & Inputs & AU & EU & CRPS & RMSE & \shortstack{Coverage\\(\%)} & Width & NLL & \shortstack{TU/\\MSE} \\
\hline
\multirow{3}{*}{\shortstack[l]{Deep\\ensemble}} & Power: 100 min & $3.108\pm 0.207$ & $1.161\pm 0.524$ & $0.0218\pm 0.0003$ & $0.0556\pm 0.0001$ & $95.5\pm 0.5$ & $0.1509\pm 0.0080$ & $-2.09\pm 0.03$ & $1.042$ \\
 & Power: 6 h & $3.108\pm 0.077$ & $1.726\pm 0.646$ & $0.0218\pm 0.0001$ & $0.0558\pm 0.0002$ & $94.7\pm 0.3$ & $0.1494\pm 0.0040$ & $-2.16\pm 0.02$ & $1.054$ \\
 & Power + speed & $1.162\pm 0.177$ & $0.784\pm 0.190$ & $0.0092\pm 0.0002$ & $0.0275\pm 0.0003$ & $95.5\pm 0.3$ & $0.0738\pm 0.0034$ & $-3.21\pm 0.07$ & $1.646$ \\
\hline
\multirow{3}{*}{\shortstack[l]{MC-\\DropConnect}} & Power: 100 min & $3.334\pm 0.147$ & $3.043\pm 0.210$ & $0.0231\pm 0.0004$ & $0.0567\pm 0.0004$ & $95.5\pm 0.3$ & $0.1652\pm 0.0044$ & $-2.00\pm 0.06$ & $1.130$ \\
 & Power: 6 h & $3.782\pm 0.896$ & $3.909\pm 0.508$ & $0.0237\pm 0.0009$ & $0.0576\pm 0.0005$ & $94.8\pm 0.8$ & $0.1717\pm 0.0178$ & $-2.03\pm 0.05$ & $1.258$ \\
 & Power + speed & $1.722\pm 0.519$ & $3.497\pm 0.593$ & $0.0125\pm 0.0008$ & $0.0331\pm 0.0028$ & $96.9\pm 0.7$ & $0.1080\pm 0.0107$ & $-2.75\pm 0.08$ & $1.885$ \\
\hline
\multirow{3}{*}{BBB} & Power: 100 min & $3.104\pm 0.210$ & $0.100\pm 0.086$ & $0.0217\pm 0.0001$ & $0.0556\pm 0.0003$ & $94.8\pm 0.6$ & $0.1441\pm 0.0042$ & $-2.09\pm 0.04$ & $1.007$ \\
 & Power: 6 h & $3.445\pm 0.489$ & $0.083\pm 0.026$ & $0.0217\pm 0.0002$ & $0.0558\pm 0.0002$ & $93.8\pm 0.6$ & $0.1429\pm 0.0074$ & $-2.08\pm 0.05$ & $1.109$ \\
 & Power + speed & $1.268\pm 0.272$ & $0.060\pm 0.010$ & $0.0110\pm 0.0006$ & $0.0303\pm 0.0006$ & $91.4\pm 1.4$ & $0.0720\pm 0.0078$ & $-3.00\pm 0.10$ & $1.390$ \\
\hline
\end{tabular}}
\end{minipage}}
\end{tablehere}
\end{minipage}\par
\begin{multicols}{2}
\raggedcolumns

\noindent\textbf{Operating-condition analysis.} The shape of the turbine power curve helps explain where wind-speed measurements improve the forecast. Between 6 and 12~m/s, a small difference in wind speed corresponds to a large difference in typical power output (Fig.~\ref{fig:scada-maps}). Knowing the current and recent wind speed should therefore add more information to the power history here than on the plateau, where typical output changes little with speed. The forecast errors follow this pattern (Fig.~\ref{fig:scada-operating}). The 6--12~m/s range contains 38\% of the test cases but accounts for 82--88\% of the reduction in squared error when wind speed is added, depending on the method. At 15--20~m/s, RMSE changes little for any method.

\end{multicols}
\noindent\begin{minipage}{\textwidth}
\begin{figurehere}
    \centering
    \includegraphics[width=0.94\textwidth]{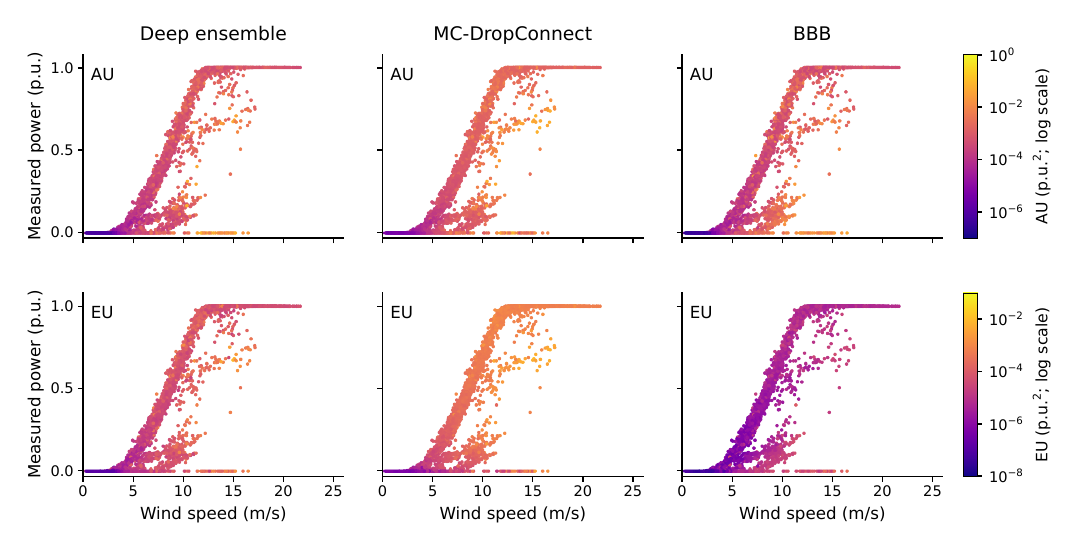}
    \caption{AU and EU across SCADA operating conditions. Points locate the 4,979 test observations by measured wind speed and power. Colors show component variances averaged pointwise over five seeds, using all training data and wind-enhanced inputs. Each row shares a logarithmic color scale.}
    \label{fig:scada-maps}
\end{figurehere}
\end{minipage}\par
\begin{multicols}{2}
\raggedcolumns

Wind speed alone also leaves some operating conditions ambiguous. At 10--15~m/s, for example, the turbine sometimes records zero power at wind speeds commonly associated with high output (Fig.~\ref{fig:scada-maps}). Even with wind speed included, all three methods generally assign higher AU to these zero-output cases than to the high-output cases. Do their predicted means also disagree more? For the ensemble and BBB they do, whereas MC-DropConnect shows greater disagreement at high output. An observation far from the main power curve can thus correspond to greater estimated variability within each model without implying greater disagreement across models.

\end{multicols}
\noindent\begin{minipage}{\textwidth}
\begin{figurehere}
    \centering
    \includegraphics[width=\textwidth]{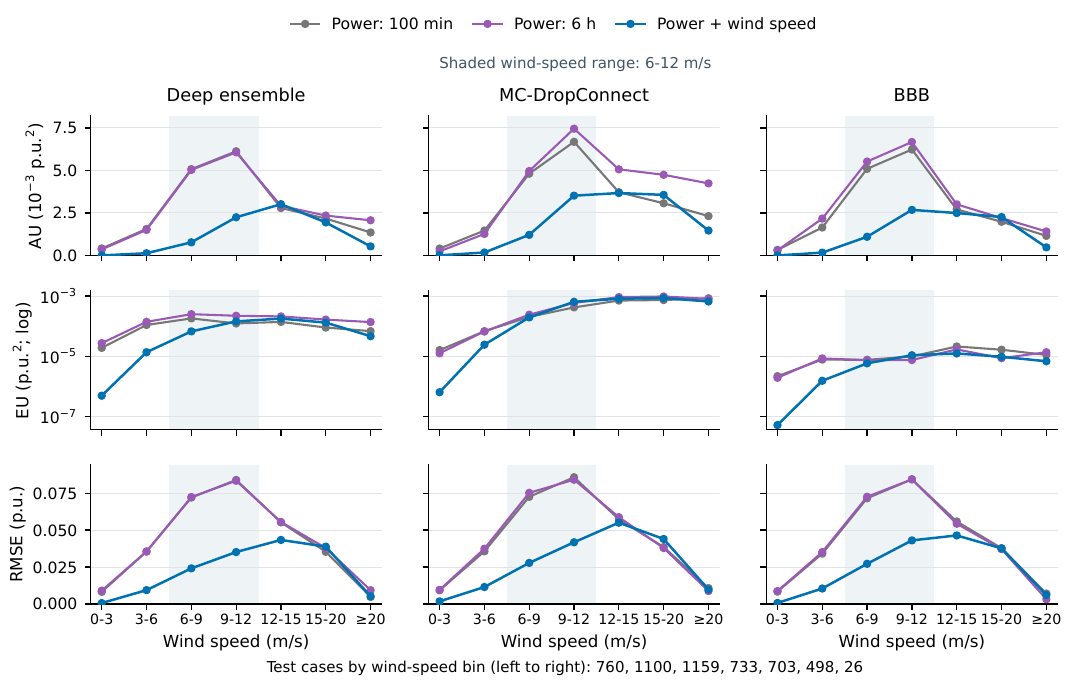}
    \caption{SCADA results by wind speed with all training data. Columns show methods; rows show mean AU, mean EU, and RMSE. Curves are five-seed averages. Shading marks 6--12 m/s; test counts are listed below in wind-bin order.}
    \label{fig:scada-operating}
\end{figurehere}
\end{minipage}\par
\begin{multicols}{2}
\raggedcolumns

\noindent\textbf{Joint interpretation.} With the forecasting inputs fixed, the deep ensemble produced wider prediction intervals at the full training share than at 10\%, even though its members agreed more closely on mean power. The average variance around the members' own predictions more than doubled, outweighing the decline in EU and increasing total predictive variance (Table~\ref{tab:scada-r1}). With the training records fixed, adding wind speed reduced estimated AU for all three methods. Their agreement on mean power responded differently: MC-DropConnect's sampled predictions showed slightly greater disagreement on average (Table~\ref{tab:scada-r2}).

\section{Conclusion}
\label{sec:3}
Wind-UDA provides a framework for defining, estimating, and assessing aleatoric and epistemic uncertainty in wind power forecasting. Its proper scoring rule formulation connects entropy- and variance-based decompositions and distinguishes the roles of training information and forecasting inputs. Additional training records help estimate the relationship between the available inputs and power. An additional measurement at prediction time can explain differences in power among cases that appeared similar under the original inputs.

In the synthetic experiments, models could agree closely on an inaccurate mean and attribute part of their shared error to outcome variability. In the turbine study, more training records improved the deep ensemble's mean power forecast and reduced disagreement among its members, even as its estimated AU increased. Adding wind speed to power history also improved forecast accuracy, but in this case all three methods estimated lower AU.

Future work could examine whether AU and EU estimates can help wind farm operators anticipate the value of new information. The goal would be to identify which additional measurements would most improve forecasts and which operating conditions need better representation in the training data.

\nonumsection{Acknowledgments}
This work was supported by the Australian Research Council (ARC), IC210100021.

\bibliographystyle{ws-jdec}
\bibliography{refer}
\end{multicols}

\end{document}